\documentclass[sigconf]{acmart}
\usepackage[utf8]{inputenc}
\usepackage[english]{babel}
\usepackage{xcolor,colortbl}
\usepackage{graphicx}
\usepackage{bm}
\usepackage{enumitem}
\usepackage{multirow}

\usepackage{makecell}
\usepackage{booktabs} 
\usepackage{caption}
\usepackage{subcaption}
\usepackage{amsmath}

\definecolor{Gray}{gray}{0.85}
\newcommand{\independent}{\perp \!\!\! \perp}

\newcolumntype{a}{>{\columncolor{Gray}}c}
\newcolumntype{b}{>{\columncolor{white}}c}
\usepackage{footnote}
\makesavenoteenv{tabular}
\usepackage[font=small]{caption}
\usepackage{dsfont}
\newtheorem{assumption}{Assumption}
\newtheorem{theorem}{Theorem}
\newtheorem{definition}{Definition}
\newcommand{\mat}[1]{{\bf #1}}   
\definecolor{ForestGreen}{rgb}{0.0, 0.27, 0.13}

\AtBeginDocument{%
  \providecommand\BibTeX{{%
    \normalfont B\kern-0.5em{\scshape i\kern-0.25em b}\kern-0.8em\TeX}}}

\setcopyright{acmcopyright}
\copyrightyear{2022}
\acmYear{2022}
\setcopyright{acmcopyright}\acmConference[WSDM '22]{Proceedings of the Fifteenth ACM International Conference on Web Search and Data Mining}{February 21--25, 2022}{Tempe, AZ, USA}
\acmBooktitle{Proceedings of the Fifteenth ACM International Conference on Web Search and Data Mining (WSDM '22), February 21--25, 2022, Tempe, AZ, USA}
\acmPrice{15.00}
\acmDOI{10.1145/3488560.3498372}
\acmISBN{978-1-4503-9132-0/22/02}
\begin{document}
\fancyhead{}
\title[Estimating Effects of Online Reviews with Multi-Modal Proxies]{Estimating Causal Effects of Multi-Aspect Online Reviews with Multi-Modal Proxies}

\author{Lu Cheng\textsuperscript{\rm 1},  Ruocheng Guo\textsuperscript{\rm 2}, Huan Liu\textsuperscript{\rm 1}}
\affiliation{\textsuperscript{\rm 1} School of Computing and Augmented Intelligence, Arizona State University, USA\\
\textsuperscript{\rm 2} School of Data Science, City University of Hong Kong, China}
\email{{lcheng35, huanliu}@asu.edu, ruocheng.guo@cityu.edu.hk}

\renewcommand{\shortauthors}{Cheng, et al.}

\begin{abstract}
Online reviews enable consumers to engage with companies and provide important feedback. Due to the complexity of the high-dimensional text, these reviews are often simplified as a single numerical score, e.g., ratings or sentiment scores. This work empirically examines the \textit{causal effects} of user-generated online reviews on a granular level: we consider \textit{multiple aspects}, e.g., the Food and Service of a restaurant. Understanding consumers' opinions toward different aspects can help evaluate business performance in detail and strategize business operations effectively. Specifically, we aim to answer \textit{interventional} questions such as \textit{What will the restaurant popularity be if the quality w.r.t. its aspect Service is increased by 10\%?} The defining challenge of causal inference with observational data is the presence of ``confounder'', which might not be observed or measured, e.g., consumers' preference to food type, rendering the estimated effects biased and high-variance. To address this challenge, we have recourse to the multi-modal proxies such as the consumer profile information and interactions between consumers and businesses. We show how to effectively leverage the rich information to identify and estimate causal effects of multiple aspects embedded in online reviews. Empirical evaluations on synthetic and real-world data corroborate the efficacy and shed light on the actionable insight of the proposed approach.
\end{abstract}



\maketitle

\section{Introduction}
\begin{figure}
\centering
  \includegraphics[width=.6\columnwidth]{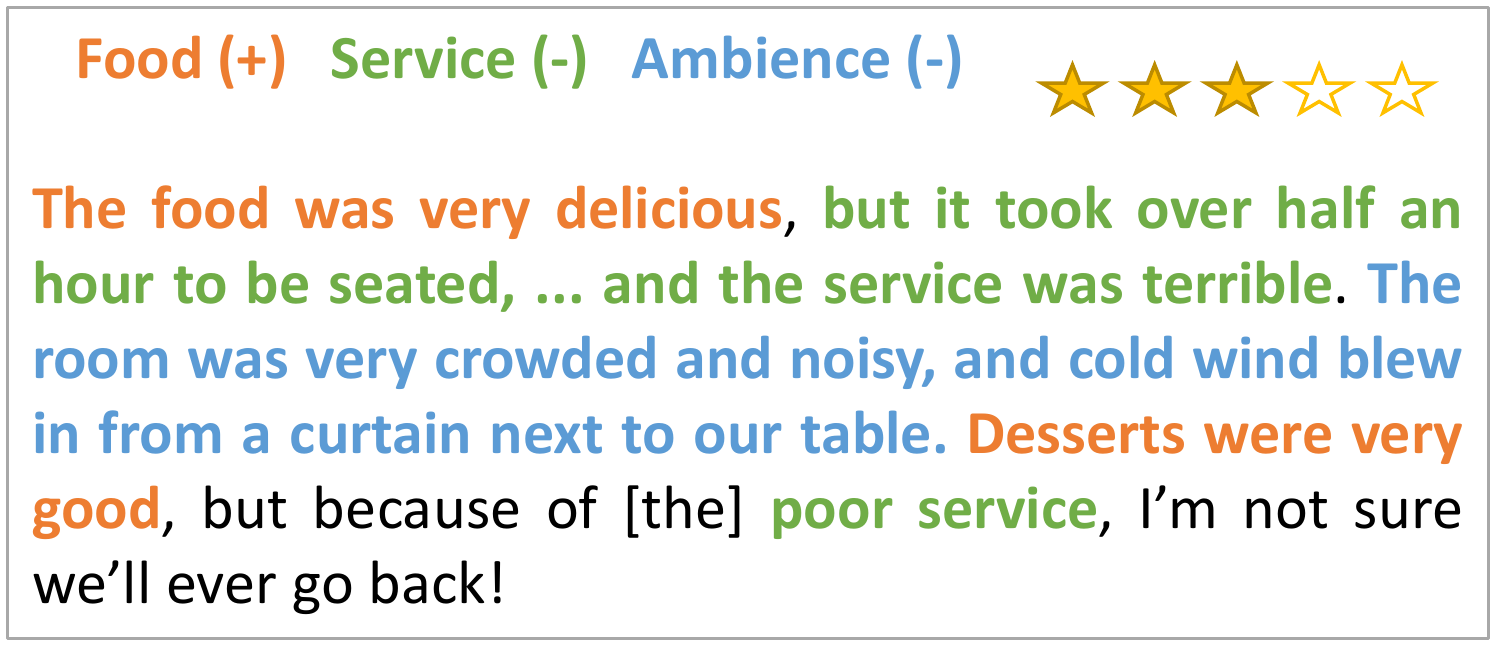}
  \caption{An example of Yelp reviews. Different aspects (e.g., Food) are highlighted in different colors. Best viewed in color.}
 \label{example}
\end{figure}

Online reviews have become a critical source to evaluate business performance, raising considerable research efforts to understanding the effects of online reviews. Due to the complexity of using the high-dimensional text in causal studies, these reviews are typically simplified as an overall numerical rating \cite{anderson2012learning} or an aggregated sentiment score \cite{wang2017impact}. However, such coarse-grained analysis provides limited information about strategizing business operations partly because it often results in findings that are inconsistent or even conflict with each other \cite{li2019effect}. Text in online reviews provides detailed information about consumers' opinions toward different aspects of a business. For example, Fig.  \ref{example} depicts a Yelp review with an overall rating of 3-star. In this figure, a plain-text review expresses a positive opinion toward the aspect food and negative opinions toward aspect ambience and service. This granular-level analysis of the content in online reviews help evaluate business performance in different dimensions and strategize operations effectively. Therefore, we seek an \textbf{aspect-level effects estimation}, a fine-grained analysis of how consumers' opinions (sentiments) toward different aspects of a business expressed in the reviews -- \textit{multi-aspect sentiment (MAS)} -- influence business popularity. 

Estimating causal effects of MAS on popularity is essentially to intervene on individual sentiment aspects and answer questions such as ``\textit{What will the restaurant popularity be if the quality w.r.t. aspect Service is increased by 10\%?}''\footnote{We assume that consumers' opinions toward an aspect reflect its quality.} Interventions could be formally justified by controlling \textit{confounding bias} -- the influence of variables that cause spurious correlations between treatment assignment\footnote{We use treatments and (potential) causes exchangeably.} (e.g., MAS) and the outcome (e.g., popularity). In observational studies, confounders might be unobserved or unmeasured, or \textit{hidden confounders} (HC), such as a consumer's food preference. Common solution is either assuming away the presence of HC or using proxy variables (i.e., observed covariates) to approximate them \cite{louizos2017causal,miao2018identifying}.

However, when applied to our problem, selecting the `right' proxies is especially challenging due to the multi-modality of online review systems such as the consumers' profile information, businesses' attributes, and consumer-restaurant interactions (i.e., a consumer writes reviews for specific restaurants). First, existing proxy strategies might lead us to include covariates insufficient to adjust for confounding biases. Second, even with sufficiently rich covariate set, one may confront the obstacle of how to exclude `bad' covariates, i.e., variables that induce bias when they are used as proxy variables. When estimating effects of online reviews on popularity, business revenue would be a `bad' proxy because it is causally affected by the outcome (i.e., popularity). These methods and their challenges, however, suggest a new approach for causal effect estimation: we might have recourse to a \textit{representation} of the \textit{multi-modal proxy variables} observed in online review systems. The intuition is that the richer the covariate set is, the more likely it is to accurately predict the outcomes and estimate the effects \cite{d2021overlap}. Previous findings (e.g., \cite{guo2020learning}) also advocated to use network information such as social networks that embed the homophily effect to learn HC. Further, learning a representation of the multi-modal covariate set rather than directly using the covariate set itself can help block the undesired biases induced by bad covariates when they are conditioned on. Therefore, desired representations should contain sufficient information for confounding adjustment, exclude biases induced by bad proxy variables, and helps provide less biased and low-variance causal estimates. This work presents a method to simultaneously learn such representations and estimate causal effects of multi-aspect online reviews on popularity.

\begin{figure}
\centering
  \includegraphics[width=\columnwidth]{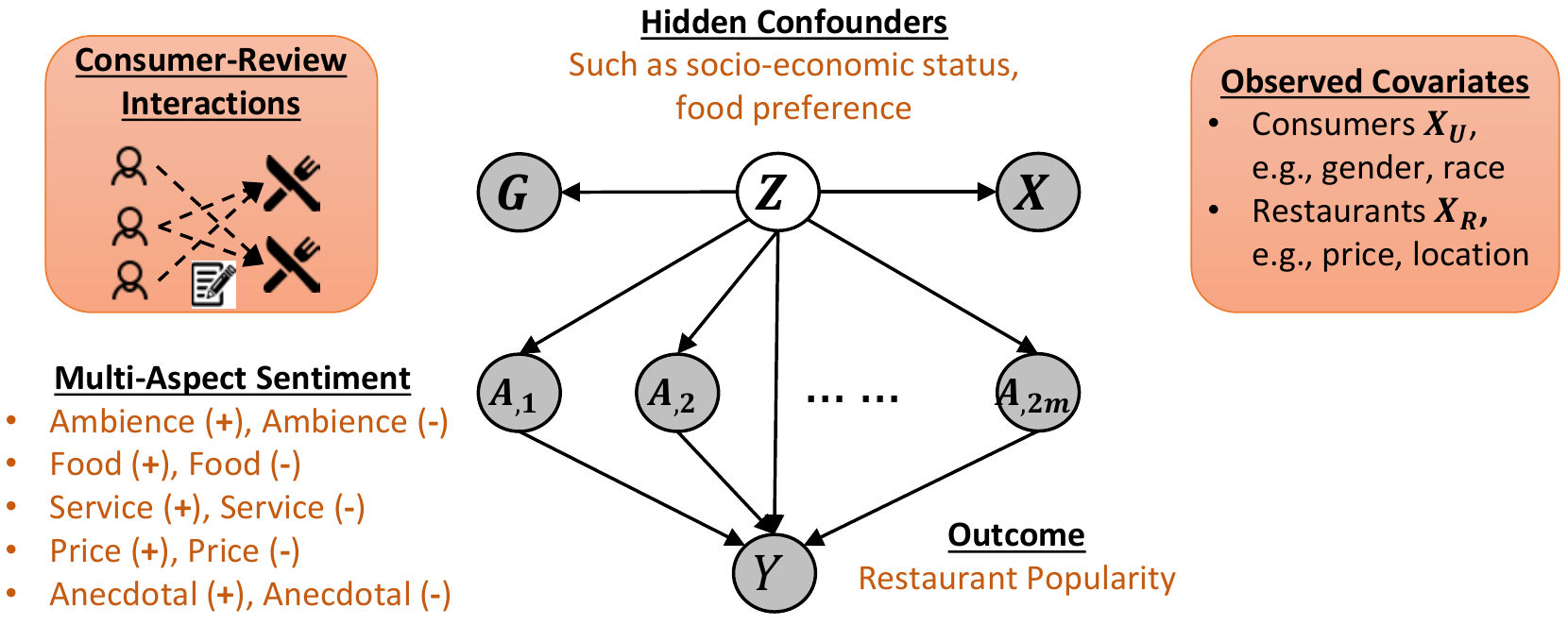}
  \caption{The causal graph for estimating effects of multi-aspect sentiment ($\mat{A_{,1}},...,\mat{A_{,2m}}$) in online reviews on restaurant popularity $Y$. We consider the presence of shared HC $\mat{Z}$ that can be approximated by multi-modal proxies (orange rectangles) and appropriate causal adjustment. The multi-modal proxies include covariates of consumers $\mat{X}_U$ and restaurants $\mat{X}_R$ as well as consumer-restaurant interactions $\mat{G}$. Individual sentiment aspect is assumed to be independent of each other given $\mat{Z}$. Best viewed in color.}
 \label{problem}
\end{figure}

\noindent\textbf{Contributions.} We develop a principled framework--\underline{D}econfounded \underline{M}ulti-Modal \underline{C}ausal \underline{E}ffect \underline{E}stimation (DMCEE) to estimate causal effects of MAS on business popularity in the presence of hidden confounding,\textit{ with low bias and variance}. DMCEE provides valid causal inferences through two primary components: (1) the \textit{Proxies Encoding Network} that maps the multi-modal proxies to a shared latent space; and (2) the \textit{Causal Adjustment Network} that seeks to extract information from the latent representation that is sufficient to adjust for the confounding while excluding undesired biases. At its core, DMCEE jointly recovers HC from observed multi-modal proxies and estimate effects of online reviews from different aspects. 

The main contributions of this work are: (1) we study a novel and practical problem to examine the effects of multiple aspects of a business expressed in online reviews on business popularity in the presence of HC (Section 2); (2) we theoretically prove that the studied problem is causally identifiable (Appendix A) and propose an effective approach that aims to estimate causal effects and approximate HC by leveraging multi-modal proxies and appropriate causal adjustment (Section 3); (3) we validate the effectiveness of DMCEE using synthetic data and two novel datasets collected from the real world (Section 4). The case studies further illustrate how DMCEE can help business owners strategize business operations.
\section{Problem Definition}
We first illustrate the problem using a causal graph described in Fig.  \ref{problem}. Given online reviews and restaurant popularity, our goal is to estimate the effects of MAS ($\mat{A}$) on popularity ($Y$) in the presence of shared HC ($\mat{Z}$). We also observe rich covariates with multi-modalities, including consumers' profile information ($\mat{X}_U$, e.g., race), restaurants' attributes ($\mat{X}_R$, e.g., price range), and consumer-restaurant interactions ($\mat{G}$) that further manifest the behavioral/operational similarity between consumers/restaurants \cite{wang2019neural}.

We now formally define the proposed problem within the standard Potential Outcome framework \cite{rubin2005causal}. Given a set of $N_R$ restaurants $\mathcal{R}$ along with a corpus of online reviews $\mathcal{C}$ written by a group of $N_U$ consumers $\mathcal{U}$, for each restaurant $r\in\{1,2,...,N_R\}$, we extract $m$ aspects from $\mathcal{C}$ and each aspect is associated with continuous positive and negative sentiment scores. Previous findings showed that positive and negative online reviews exert different influences on the outcomes \cite{tsao2019asymmetric}. We denote the MAS for restaurant $r$ as a $2m$-dimensional vector $\bm{a}_r=<a_{r1},a_{r2},...,a_{r2m}>$. A \textit{potential outcome function} then maps $\bm{a}_r$ to the outcome $y_r(\bm{a}_r)$, i.e., $y_r(\bm{a}_r): \mathbb{R}^{2m} \rightarrow \mathbb{R}.$ The fundamental problem in causal inference is that only one potential outcome can be observed \cite{guo2020survey}, e.g., $Y_r(\mat{A}_r)$ for a certain configuration of $\mat{A}_r$. Let $\mat{X}_R$ and $\mat{X}_U$ represent restaurant and consumer covariates, respectively. Interactions between consumers and restaurants are defined as a bipartite graph $\mathcal{G}=(\mathcal{U},\mathcal{R},\mathcal{E})$ where nodes consist of consumers and restaurants, and edges set $\mathcal{E}$ denotes review-writing relationships. The adjacency matrix of $\mathcal{G}$ is $\mat{G}\in \mathbb{R}^{N_R\times N_U}$ where 1 denotes $u\in\mathcal{U}$ writes a review for $r\in\mathcal{R}$, 0 otherwise. 

We now define the problem as follows:
\begin{definition}[Estimating Effects of Multi-Aspect Online Reviews] Given the multi-modal information, we aim to study how sentiment aspects in $\mat{A}\in\mathcal{A}$ extracted from $\mathcal{C}$ influence the restaurant popularity $Y$ in the presence of HC $\mat{Z}$. Specifically, we are interested in jointly estimating the average treatment effect (ATE) of individual sentiment aspect $\mat{A}_{,j},j\in\{1,2,..,2m\}$ on $Y$:
\begin{equation}
     \tau_j=\mathbb{E}[Y_r(a_{rj})]-\mathbb{E}[Y_r(a'_{rj})]\quad r\in\{1,2,...,N_R\},
\end{equation}
where $\mathbb{E}[Y_r(a_{rj})]$ denotes the expected restaurant popularity with aspect sentiment score $a_{rj}$.
\end{definition}
\noindent As the aspect sentiment scores are continuous and there are multiple aspects, this is essentially a continuous multiple treatment (dose) effect estimation problem. The estimation is unbiased iff $\mathbb{E}[Y_r(a_{rj})]=\mathbb{E}[Y_r(a_{rj})|A_{rj}=a_{rj}]$, i.e., all confounding is properly adjusted for. HC can generate a statistical dependency between $\mat{A}$ and $Y$, i.e., $\mathbb{E}[Y_r(a_{rj})]\neq\mathbb{E}[Y_r(a_{rj})|A_{rj}=a_{rj}]$. Our goal is to alleviate confounding when inferring the expected popularity $\mathbb{E}[Y_r(a_{rj})]$.
\section{Causal Estimation with Multi-Modal Proxies}
\begin{figure}
\centering
  \includegraphics[width=.8\columnwidth]{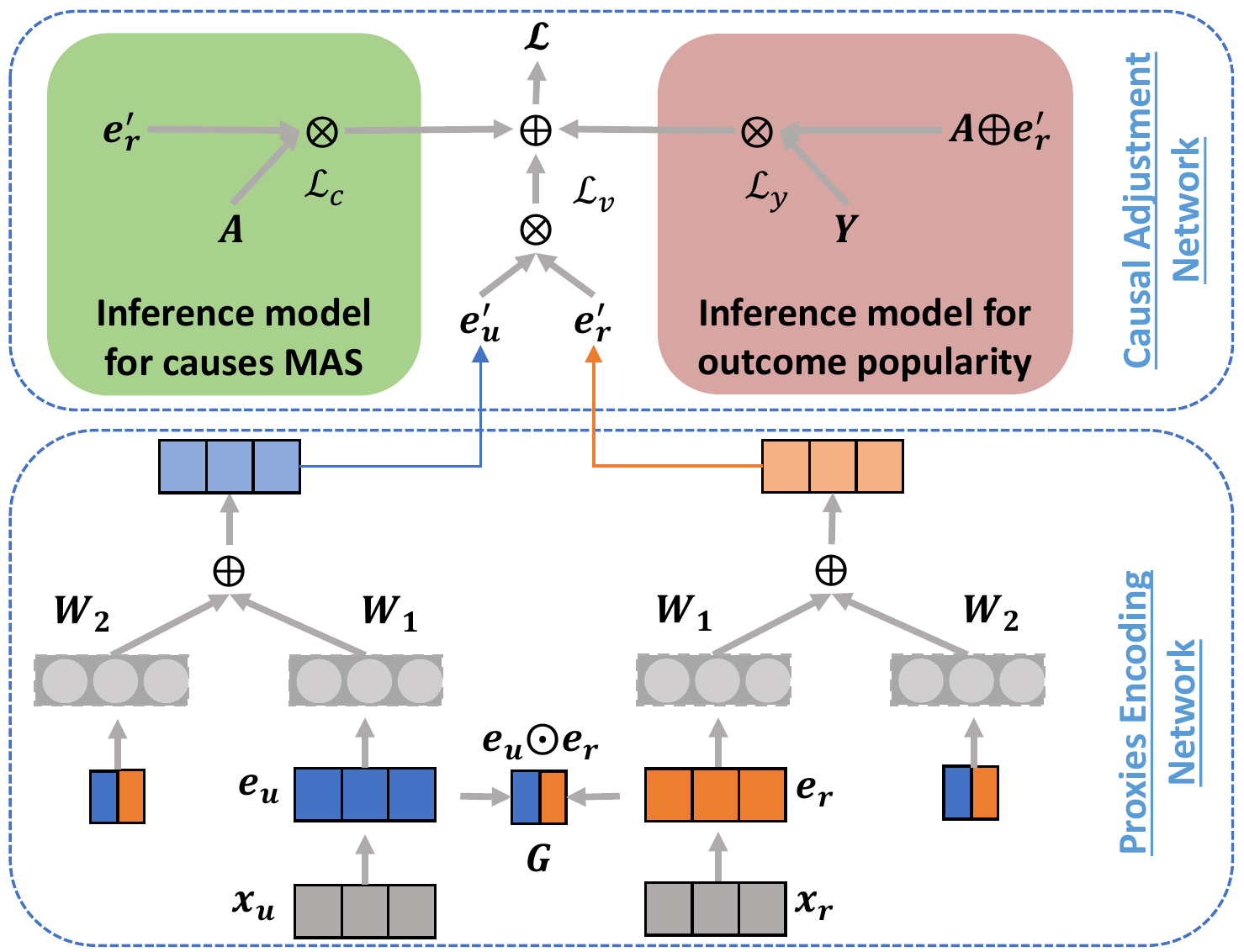}
  \caption{An overview of DMCEE framework. The proxies encoding network (bottom) leverages multi-modal proxies to encode the consumer and restaurant representations ($\mat{e}'_u, \mat{e}'_r$). The causal adjustment network (top) then extracts information from the representations to control for confounding while excluding undesired biases. The final loss ($\mathcal{L}$) consists of the reconstruction error ($\mathcal{L}_v$), the MAS ($\mathcal{L}_C$) and the outcome ($\mathcal{L}_y$) prediction errors.}
 \label{framework}
\end{figure}

There have been many discussions about the validity of using proxy strategies to adjust for HC \cite{miao2018identifying}. For multi-modal proxies, we confront two primary challenges: first, only including simple covariates may be insufficient to control for confounding biases in the observational data; second, if the covariates are sufficiently rich, we may include variables that induce bias when they are conditioned on, i.e., the `bad controls' such as restaurant revenues. In this work, we therefore propose to adjust for the representation of multi-modal proxies by following the causal mechanism illustrated in Fig. \ref{problem}, rather than the covariates themselves. DMCEE consists of two major components: (1) the proxies encoding network that constructs a representation in the latent space using multi-modal proxies; and (2) the causal adjustment network that extracts sufficient causal information from the representation to account for confounding while excluding the induced bias from the bad controls. Fig.  \ref{framework} features the overview framework of DMCEE. Discussion of the causal identification of DMCEE can be found in Appendix A. 
\subsection{Proxies Encoding Network}
Multi-modal proxies in online review systems can be implicitly connected, e.g., the consumer-restaurant interaction $\mat{G}$, therefore, violating the standard i.i.d. assumption. Representation learning offers an effective alternative to fully specify the joint probability distributions of observed data with non-i.i.d variables \cite{pmlr-v124-veitch20a}. Particularly, we leverage the Graph Convolution Network (GCN) \cite{hamilton2017inductive,kipf2016semi} -- a powerful neural network that efficiently produces useful feature representations of nodes in the network \cite{berg2017graph} -- to construct the representations of consumers and restaurants using the multi-modal covariate set $\{\mat{X}_U, \mat{X}_R, \mat{G}\}$. 
Proxies encoding network consists of two steps: message construction and message aggregation. 
\subsubsection{Message Construction} 
Given restaurant $r\in\mathcal{R}$ (consumer $u\in\mathcal{U}$) with covariates $\mat{x}_{r} \in \mathbb{R}^{d_r}$ ($\mat{x}_{u}\in \mathbb{R}^{d_u}$), where $d_r$ ($d_u$) denotes the covariate dimension, we first transform $\mat{x}_{r}$ and $\mat{x}_{u}$ into a latent space via two weight matrices\footnote{We omit the bias terms here for simplicity.} $\mat{W}_R \in \mathbb{R}^{d_r\times D}$ and $\mat{W}_U\in \mathbb{R}^{d_u\times D}$:
\begin{gather}
    \mat{x}_{r} \rightarrow \mat{e}_{r}: \mat{e}_{r}=\mat{x}_{r}\mat{W}_R;\quad
     \mat{x}_{u} \rightarrow \mat{e}_{u}: \mat{e}_{u}=\mat{x}_{u}\mat{W}_U.
\end{gather}
$\mat{e}_{r}\in\mathbb{R}^D$ $(\mat{e}_{u}\in \mathbb{R}^D)$ is the restaurant (consumer) representation with $D$ denoting the size of hidden units. Intuitively, the consumer-restaurant interactions $\mat{G}$ contains hidden information about restaurant attributes and consumer behaviors \cite{guo2020learning}. For example, restaurants that are often visited by the same group of consumers may provide information about the shared consumers' dining habits. We thereby include $\mat{G}$ as a proxy variable of HC and incorporate it into the encoding procedure. Given a pair $(u,r)$, we define the representation of message from $u$ to $r$ as:
\begin{equation}
    \mat{m}_{r\leftarrow u}=p_{ru}(\mat{W}_1\mat{e}_u+\mat{W}_2(\mat{e}_u\odot\mat{e}_r)); \quad p_{ru}=\frac{1}{\sqrt{|\mathcal{N}_r||\mathcal{N}_u|}},
\end{equation}
where $\mat{W}_1$, $\mat{W}_2 \in \mathbb{R}^{d'\times D}$ are the trainable weight matrices and $d'$ is the number of hidden units. The element-wise product $\odot$ between $\mat{e}_r$ and $\mat{e}_u$ effectively encodes the consumer-restaurant interactions as the operation passes more messages from the similar consumers \cite{wang2019neural}. $p_{ru}$ is defined in \cite{kipf2016semi}, where $\mathcal{N}_r$ and $\mathcal{N}_u$ denote one-hop neighbors of restaurant $r$ and consumer $u$. $p_{ru}$ can be interpreted as the contribution from consumers to inferring the HC. 
\subsubsection{Message Aggregation} 
Next, we aggregate the messages propagated from $r$'s neighbors to further refine $\mat{e}_r$:
\begin{equation}
    \mat{e}'_r=\text{LeakyReLU}(\mat{m}_{r\leftarrow r}+\sum_{u\in \mathcal{N}_r}\mat{m}_{r\leftarrow u}),
\end{equation}
where LeakyReLU \cite{maas2013rectifier} is the activation function. As advocated in \cite{wang2019neural}, we also retain $r$'s original covariates by considering the self-connection of $r$: $\mat{m}_{r\leftarrow r}=\mat{W}_1\mat{e}_r$. Analogously, we can obtain $\mat{e}'_u$ of consumer $u$. Optimization is achieved by predicting the consumer's tendency towards visiting a particular restaurant. The loss function of the proxies encoding network is defined as:
\begin{gather}
    \mathcal{L}_v=\sum_{(u,i,j)\in\mathcal{O}}-\ln\sigma(\hat{v}_{ui}-\hat{v}_{uj}),\quad\hat{v}(u,r)=\mat{e}_u^{'\intercal}\mat{e}'_r,
\end{gather}
where the inner product encodes the consumer-restaurant interactions, $\mathcal{O}=\{(u,i,j)|(u,i)\in\mathcal{E}^+,(u,j)\in \mathcal{E}^-\}$ denotes the pairwise training data. $\mathcal{E}^+$ and $\mathcal{E}^-$ indicate the observed and unobserved consumer-restaurant interactions. $\sigma(\cdot)$ is the sigmoid function.
\subsection{Causal Adjustment Network}
With the encoded consumer ($\mat{e}'_U$) and restaurant ($\mat{e}'_R$) representations, the causal adjustment network learns a surrogate confounder that aim to exclude the biases induced by potentially `bad' proxy variables while ensuring to contain sufficient information for the confounding adjustment. 

We start with a direct corollary of Thm.3 in \cite{rosenbaum1983central}:
\begin{theorem} Suppose $\gamma(\mat{Z})$ is some function of HC $\mat{Z}$ such that at least one of the following is $\gamma(\mat{Z}$)-measurable: (i) $P(Y|\mat{A},\mat{Z})$, or (ii) $P(\mat{A}|\mat{Z})$. If $\mat{Z}$ suffices to fully block the backdoor paths between MAS $\mat{A}$ and popularity $Y$, then $\gamma(\mat{Z})$ also suffices. 
\end{theorem}
\noindent Thm. 1 indicates that, to alleviate the confounding bias, we do not need to fully recover $\mat{Z}$ but the aspects $\gamma(\mat{Z})$ that suffice to accurately predict treatment assignment and the outcome \cite{pmlr-v124-veitch20a}. 

Under Thm. 1, the causal adjustment network seeks a function $\gamma(\mat{e}'_R)$ such that the output representation $\bm{\gamma}_R$ can be directly used as a surrogate confounder to estimate the effects of MAS with low bias and variance. In particular, we build two inference models for the causes MAS and outcome popularity that map (1) $\mat{e}'_R$ to predictions for MAS ($\mat{A}$); (2) $\mat{e}'_R$ and $\mat{A}$ to predictions for restaurant popularity ($Y$). We first introduce the underlying assumptions of the causal adjustment network as shown below:
\begin{assumption}[Estimation Assumptions] \
\begin{enumerate}
    \item There is a function $\gamma(\cdot)$ such that it takes the input of $\mat{e}'_R$ and outputs representation $\bm{\gamma}_R$ that blocks all backdoor paths from multiple causes $\mat{A}$ to outcome $Y$, i.e., $\mat{A}\independent Y|\bm{\gamma}_R$;
    \item The probability of observing $\mat{A}$ given any feasible $\bm{\gamma}_R$ is positive, i.e., $P(\mat{A}|\bm{\gamma}_R)>0,\ \mat{A}\in \mathcal{A}$. 
    \item The conditional outcome model $\hat{\mat{y}}=P(\mat{y}|\mat{A},\bm{\gamma}_R)$ is consistent, i.e., $|\mat{y}-\hat{\mat{y}}|\rightarrow 0$ as $N_R\rightarrow \infty$. 
\end{enumerate}
\end{assumption}  
\noindent  Asm. 1.1 implies that the surrogate confounder $\bm{\gamma}_R$ is shared across all sentiment aspects and it can fully block the backdoor paths (i.e., capture the confounding bias) between MAS and the popularity. Asm. 1.2-1.3 are standard assumptions (i.e., positivity and consistency) in causal effect estimation. Asm. 1.2 requires that the probability of observing a sentiment aspect given $\bm{\gamma}_R$ is positive and Asm. 1.3 indicates that the potential popularity of a restaurant inferred from $\mat{A}$ and $\bm{\gamma}_R$ is precisely its observed popularity.  

Suppose we are interested in the restaurant popularity during a specific time period\footnote{It is possible to consider the distribution of popularity over a week as the outcome, however, this requires more advanced approaches that are left as a future work.} (e.g., 07:00 PM - 08:00 PM on Saturday), denoted as $Y$. Our goal is to jointly estimate ATEs of individual sentiment aspects on $Y$. Thm. 1 reduces the causal adjustment task to joint predictions of the causes MAS and outcome popularity. We first define the inference models of outcome and causes as
\begin{equation}
\begin{gathered}
    Q(\bm{a},\bm{\gamma}_r)=\mathbb{E}[Y_r|\mat{A}_r=\bm{a}, \gamma(\mat{e}'_r)=\bm{\gamma}_r];\quad q(\bm{\gamma}_r)=P(\mat{A}_r|\gamma(\mat{e}'_r)=\bm{\gamma}_r).
\end{gathered}
\end{equation}
To illustrate, $q(\bm{\gamma}_r)$ can denote the probability of restaurant $r$ receiving sentiment score 0.8 w.r.t. aspect Food given $\bm{\gamma}_r$. 
Here, we replace proxies $\mat{X}$ and $\mat{G}$ with $\bm{\gamma}_r$ as $\bm{\gamma}_r$ represents information in $\mat{X}$ and $\mat{G}$ that suffices to estimate the ATE of individual sentiment aspect, i.e., decouple the spurious correlation between $\mat{A}$ and $Y$. 
The objective of predicting MAS can be defined as:
\begin{equation}
    \mathcal{L}_c=\frac{1}{N_R}\sum_{r=1}^{N_R}\sum_{j=1}^{2m}f_c(A_{rj},\bm{\gamma}_r;\bm{\theta_c}),
\end{equation}
where $f_c(\cdot)$ measures the prediction errors and $\bm{\theta}_c=\{\theta_{1}, \theta_{2}, ..., \theta_{2m}\}$ comprises the model parameters for MAS predictions. The inference model for popularity is 
\begin{equation}
    \mathcal{L}_y=\frac{1}{N_R}\sum_{r=1}^{N_R}f_y(y_r,\mat{A}_r,\bm{\gamma}_r;\bm{\theta}_y),
\end{equation}
where $f_y(\cdot)$ measures the differences between the true and predicted popularity, and $\bm{\theta}_y=[\bm{\theta}_A\circ\bm{\theta}_\gamma]$ denotes the model parameter that maps MAS with $\bm{\theta}_A$ and $\bm{\gamma}_r$ with $\bm{\theta_\gamma}$ to the outcome space. $\circ$ represents the operation of concatenation. As $\mat{A}$ and $Y$ are continuous, we define $f_c(\cdot)$ and $f_y(\cdot)$ as the Mean Squared Error (MSE):
\begin{equation}
    f_c(\cdot)=\big\|\bm{\gamma}\bm{\theta}_c-\mat{A}\big\|^2_2;\quad
    f_y(\cdot)=\big\|[\mat{A}\circ \bm{\gamma}]\bm{\theta}_y-\mat{y}\big\|_2^2,
\end{equation}
The final objective function of DMCEE is the weighted sum:
\begin{equation}
\begin{aligned}
\mathcal{L}=\alpha\mathcal{L}_v+\beta\mathcal{L}_c+\mathcal{L}_y+\lambda\|\Theta\|^2_2,
\end{aligned}
\end{equation}
where $\alpha$, $\beta$ are hyperparameters that balance the contribution of each module to the final loss. $\Theta$ denotes all trainable model parameters and $\lambda$ controls the $\ell_2$ regularization strength to prevent overfitting. $\bm{\hat{\theta}_A}$ represents the estimated causal effects of MAS.
\section{Experiments}
Evaluating causal inference methods with observational data has been a long-standing challenge due to the lack of ground-truth effects \cite{louizos2017causal}. Simulation therefore becomes a common alternative to evaluate the performance of causal models. In this section, we start with synthetic experiments that simulate online reviews. The second set of experiments is conducted on two real-world datasets curated by merging two independent data sources. As the causal effects are unknown in these observational studies, we follow a conventional evaluation method that performs prediction tasks on data following different distributions, a.k.a. \textit{out-of-distribution (OOD)} or \textit{invariant predictions} \cite{arjovsky2019invariant}. In particular, we are interested in answering the following research questions: 
\begin{itemize}[leftmargin=*]
    \item \textbf{RQ1.} In synthetic settings, can DMCEE improve the accuracy of estimating the HC and multi-aspect causal effects over existing methods (Section 4.2)?
    \item \textbf{RQ2.} When applied to real-world datasets with unknown causal effects, can DMCEE outperform existing approaches w.r.t. OOD popularity predictions (Section 4.3)?
    \item \textbf{RQ3.} How do the proxies encoding network and the causal adjustment network affect the results, respectively (Section 4.2-4.4)?
    \item \textbf{RQ4.} What advice and actionable insights regarding each business aspect can DMCEE provide for business owners to gain popularity (Section 4.3)?
\end{itemize}
\subsection{Experimental Setup}
Given that our problem targets multiple continuous causes, standard approaches (e.g., doubly-robust estimation \cite{bang2005doubly}) for causal effect estimation are not applicable. We thereby consider standard machine learning models, causal inference models, and a recently proposed model for multi-cause effect estimation \cite{wang2019blessings} as the baselines in the experiments. We also compare DMCEE with its two variants to examine the effectiveness of each of its network components. The details of these baselines can be seen in the following:
\begin{itemize}[leftmargin=*]
    \item \textbf{Linear Regression (LR)}: This is a non-causal model with MAS ($\mat{A}$) as the input. The learned coefficients are the estimated effects.
    \item $\mat{LR_{con}}:$ This is a causal model deconfounded by observed covariates. The learned coefficients of $\mat{A}$ are the estimated effects.
    \item $\textbf{GCN}:$ This is a causal model similar to $\mat{LR_{con}}$. But we replace covariates with learned restaurant representations.
    \item \textbf{Deconfounder (Deconf)} \cite{wang2019blessings}: This is the state-of-the-art model in multi-cause effect estimation. It leverages the latent-variable models to estimate HC by directly factorizing the causes. 
    \item $\mat{D_{Out}}$: This is a variant of DMCEE where only inference model for outcome is considered, i.e., $\beta=0$.
    \item $\mat{D_{Cau}}$: This is a variant of DMCEE where only inference model for causes is considered, i.e., $\mathcal{L}_y=0$. 
\end{itemize}
Note for $\textbf{GCN}$, \textbf{Deconfounder}, and $\mat{D_{Cau}}$, we further need to feed the approximated HC and MAS into outcome model LR\footnote{We use LR for fair comparisons and convenience to get the estimated effects.} to obtain the estimated effects. For DMCEE and its variants, unless otherwise specified, we set $\alpha=1\mathrm{e}{-}6$ and $\beta=1\mathrm{e}{-}6$ based on the parameter analysis (Detailed in Section 4.4). $\lambda$ is set to $1\mathrm{e}{-}8$. We consider Probabilistic PCA \cite{tipping1999probabilistic} as the latent variable model in Deconfounder as suggested in \cite{wang2019blessings}. The dimensions of restaurant and consumer representations are set to 15 for all models. In synthetic settings, for ATE estimation, we use the standard evaluation metric Absolute Error denoted as $\epsilon^{ATE}=|\tau_j-\hat{\tau}_j| \quad \forall j\in \{1,2,...,2m\}$.
In addition, we evaluate the quality of the learned surrogate confounder $\bm{\gamma}_R$ by computing Frobenius norm to measure the discrepancies between the approximated ($\bm{\gamma}_R$) and true ($\mat{Z}$) HC. In real-world settings, as we cannot know the ground-truth effects of individual sentiment aspects, we have recourse to OOD predictions and evaluate the accuracy of predicted outcomes using standard metrics for regression -- Mean Absolute Error (MAE) and MSE. All the best results reported in this section are statistically significant at level 0.05\footnote{The results are based on two-sided Student's t-test.}. More implementation details can be seen in Appendix B.
\begin{table}
\setlength\tabcolsep{1.5pt}
\small
\begin{center}
    \caption{Abs errors for 100 restaurants and 1000 consumers.}
    \begin{tabular}{|c|c|c|c|c|c|c|c|c|c|c|>{\columncolor[gray]{0.8}}c|}\hline
         Causes& $a_1$&$a_2$&$a_3$&$a_4$&$a_5$&$a_6$&$a_7$&$a_8$&$a_9$&$a_{10}$&Mean \\ \hline\hline
         \textbf{LR}&11.6 & 16.7 & \textbf{2.3} & 63.0 & 1.8 & 25.9 & 45.3 & 28.6 & 77.2 & 21.4 & 29.4\\ \hline
         $\mat{LR_{con}}$&12.7 & 14.1 & 24.4 & \textbf{10.7} & 1.9 & 18.9 & 26.9 & 4.2 & 34.4 & \textbf{1.9} & 15.0\\\hline
          $\mat{GCN}$&4.9 & 16.8 & 18.1 & 20.0 & 8.2 & 10.1 & \textbf{8.6} & 4.2 & 8.8 & 37.1 & 13.7 \\\hline
         \textbf{Deconf}&41.7 & \textbf{2.9} & 37.6 & 72.1 & 68.2 & 34.0 & 41.3 & 84.3 & \textbf{3.5} & 121.7 & 50.7 \\\hline
         $\mat{D_{Out}}$&\textbf{1.8} & 6.9 & 16.1 & 21.3 & \textbf{1.6} & 9.6 & 10.3 & 19.4 & 5.6 & 37.3 & 13.0 \\\hline
         $\mat{D_{Cau}}$&6.6 & 14.4 & 12.4 & 97.7 & 20.6 & 27.9 & 62.4 & 73.7 & 98.8 & 43.8 & 45.8 \\\hline
         \textbf{DMCEE}&4.0 & 14.9 & 16.7 & 16.7 & 7.7 & \textbf{8.3} & 8.7 & \textbf{3.6} & 9.0 & 34.2 & \textbf{12.4} \\\hline
    \end{tabular}
\label{syn1}
\end{center}
\end{table}
\begin{table}
\setlength\tabcolsep{1.5pt}
\small
\begin{center}
    \caption{Abs errors for 500 restaurants and 5000 consumers.}
    \begin{tabular}{|c|c|c|c|c|c|c|c|c|c|c|>{\columncolor[gray]{0.8}}c|}\hline
         Causes& $a_1$&$a_2$&$a_3$&$a_4$&$a_5$&$a_6$&$a_7$&$a_8$&$a_9$&$a_{10}$&Mean \\ \hline\hline
         \textbf{LR}&44.6 & 21.1 & 16.9 & 8.5 & \textbf{8.0} & 6.5 & 8.0 & 2.6 & 16.7 & \textbf{0.5} & 13.4\\ \hline
         $\mat{LR_{con}}$&27.7 & 19.6 & \textbf{0.9} & 0.5 & 21.9 & 7.5 & \textbf{4.8} & 0.5 & 2.9 & 8.5 & 9.5\\\hline
          $\mat{GCN}$&12.6 & 7.8 & 1.8 & 1.9 & 32.9 & \textbf{3.1} & 8.8 & \textbf{0.1} & 3.2 & 2.1 & 7.4 \\\hline
         \textbf{Deconf}&56.5 & 18.8 & 32.7 & 15.0 & 12.9 & 21.2 & 35.9 & 18.1 & 5.7 & 5.4 & 22.2 \\\hline
         $\mat{D_{Out}}$&14.4 & \textbf{6.1} & 1.1 & 9.4 & 30.9 & 8.3 & 7.4 & 0.7 & \textbf{0.8} & 3.5 & 8.2\\\hline
         $\mat{D_{Cau}}$&41.4 & 20.3 & 12.8 & 5.0 & 17.9 & 6.9 & 9.2 & 2.3 & 8.2 & \textbf{0.5} & 12.5\\\hline
         \textbf{DMCEE}&\textbf{10.4} & 9.5 & 3.8 & \textbf{0.3 }& 29.2 & 7.2 & 6.7 & 0.7 & 0.9 & 0.6 & \textbf{6.9}\\\hline
    \end{tabular}
    \label{syn2}
\end{center}
\end{table}
\begin{table}
\setlength\tabcolsep{1.5pt}
\small
\begin{center}
    \caption{Abs errors for 1000 restaurants and 10000 consumers.}
    \begin{tabular}{|c|c|c|c|c|c|c|c|c|c|c|>{\columncolor[gray]{0.8}}c|}\hline
         Causes& $a_1$&$a_2$&$a_3$&$a_4$&$a_5$&$a_6$&$a_7$&$a_8$&$a_9$&$a_{10}$&Mean \\ \hline\hline
         \textbf{LR}&16.4 & 12.4 & 14.1 & 14.9 & 11.9 & 13.7 & 25.4 & 2.9 & 42.4 & 26.9 & 18.1\\ \hline
         $\mat{LR_{con}}$&8.9 & 14.9 & 13.9 & 11.8 & \textbf{6.1} & 9.3 & 22.9 & 12.5 & 29.8 & 15.0 & 14.5 \\\hline
         $\mat{GCN}$&14.3 & \textbf{4.0} & 2.7 & 28.3 & 8.7 & 16.1 & \textbf{0.6} & 20.9 & \textbf{7.1} & 22.3 & 12.5 \\\hline
         \textbf{Deconf}&45.8 & 22.7 & 26.6 & 17.9 & 20.9 & 15.2 & 3.8 & 30.5 & 14.3 & 45.4 & 24.3\\\hline
         $\mat{D_{Out}}$&14.3 & 5.8 & \textbf{1.8} & 28.4 & 7.9 & 19.4 & 0.8 & 15.0 & 7.6 & 21.7 & 12.3\\\hline
         $\mat{D_{Cau}}$&14.2 & 8.4 & 14.3 & 15.6 & 11.8 & 14.4 & 26.5 & \textbf{0.6} & 42.3 & 26.3 & 17.4\\\hline
         \textbf{DMCEE}&\textbf{0.3} & 21.5 & 2.7 &\textbf{ 3.4 }& 16.8 & \textbf{4.4} & 5.4 & 22.2 & 11.6 & \textbf{9.5} & \textbf{9.8}\\\hline
    \end{tabular}
    \label{syn3}
\end{center}
\end{table}
\subsection{Experiments on Synthetic Online Reviews}
Under the causal mechanism described in Fig. \ref{problem}, our simulation process starts with generating multi-modal proxies including covariates of restaurants $\Tilde{\mat{R}}$ and consumers $\Tilde{\mat{U}}$. We then create the consumer-restaurant bipartite graph $\mat{G}$ based on the similarities between their covariates. Hidden confounder ($\mat{Z}$) is generated by incorporating information of $\mat{G}$, $\Tilde{\mat{R}}$, and $\Tilde{\mat{U}}$. At last, we generate causes $\mat{A}$ from $\mat{Z}$, and outcome $Y$ from $\mat{A}$ and $\mat{Z}$.
\subsubsection{Data Generation.} Specifically, the synthetic data generating process (DGP) is as follows:
\begin{equation}
\small
    \begin{gathered}
        \Tilde{\mat{R}}\sim \mathcal{N}(0,1);\quad\Tilde{\mat{U}}\sim \text{Poisson}(1);\quad
        \mat{s}=\Tilde{\mat{U}}\Tilde{\mat{R}}^T; \quad 
        \mat{G}=\text{Bern}(1,\mat{s}),\\ \mat{A}_{,j}=\sum_{u=1}^{N_U}(2\text{Bin}(1,\mat{s}_{u})\odot\mat{G}-\mat{1}) \quad \forall j\in \{1,2,...,2m\};\quad \mat{Z}=\Tilde{\mat{R}}\circ (\mat{G}^T\Tilde{\mat{U}}),\\
        \mat{W}_{\mat{A}_{,j}}=\text{Poisson}(2j) \quad\forall j;
        \mat{W}_Z=\text{Poisson}(10);\quad \mat{y}=\mat{Z}\mat{W}_Z+\mat{A}\mat{W}_A,
    \end{gathered}
\end{equation}
where $\Tilde{\mat{R}}\in \mathbb{R}^{N_R\times D}$, $\Tilde{\mat{U}}\in\mathbb{R}^{N_U\times D}$, and $\mat{W}_{\mat{A}_{,j}}$ is the $j$-th column of $\mat{W}_{A}$. Each entry in $\mat{G}\in \mathbb{R}^{N_U\times N_R}$ is sampled from Bernoulli distribution with normalized parameters dependent on $\Tilde{\mat{R}}$ and $\Tilde{\mat{U}}$. In this DGP, we assume that a consumer is more likely to write reviews for a restaurant if their covariates are more similar. The sentiment score of an aspect ($\mat{A}_{,j}$) is simulated under the assumption that the more likely a consumer will go to a restaurant, the more positive reviews this consumer will write for the restaurant, i.e., a larger sentiment aspect score will be assigned to the restaurant. We set the number of sentiment aspects to 10 (same as the real-world data, detailed in Section 4.3). The causal effects ($\mat{W}_A$) and the coefficients of HC ($ \mat{W}_Z$) are sampled from Poisson distributions with different parameters. The outcome is a linear combination of the weighted $\mat{Z}$ and $\mat{A}$. To generate the `observed' attributes for consumers and restaurants, we randomly sample half of covariates from $\Tilde{\mat{U}}$ and $\Tilde{\mat{R}}$ and denote them as $\mat{X}_U$ and $\mat{X}_R$, respectively. We consider three DGPs (the first and second entries are the number of restaurants and consumers, respectively): (100, 1000), (500, 5000), and (1000, 10000). To generate training and test datasets, we adopt $k$-means to split the dataset into $k=5$ clusters based on $\mat{X}_R$ and randomly set one cluster as the test dataset. For each DGP, we create 10 replicates and report the averaged results below.
\subsubsection{Estimating Effects of MAS} We report Absolute (Abs) Errors for estimating ATEs for individual sentiment aspects in Table \ref{syn1}-\ref{syn3}. We observe that the best performance w.r.t. each sentiment aspect is achieved by various models whereas DMCEE accomplishes the best estimations of mean ATEs of all MAS for all three DGPs, especially for DGP with smaller sample size. DMCEE also outperforms its variants $D_{Out}$ and $D_{Cau}$, revealing the importance to exclude undesired biases induced by conditioning on bad proxies. These results indicate that 1) it is challenging to simultaneously optimize effect estimations for all sentiment aspects; and 2) DMCEE can improve the effects estimation accuracy and the improvement tends to be more significant for smaller data.
\subsubsection{Hidden Confounder Estimation} 
\begin{figure}
\centering
  \includegraphics[width=.55\columnwidth]{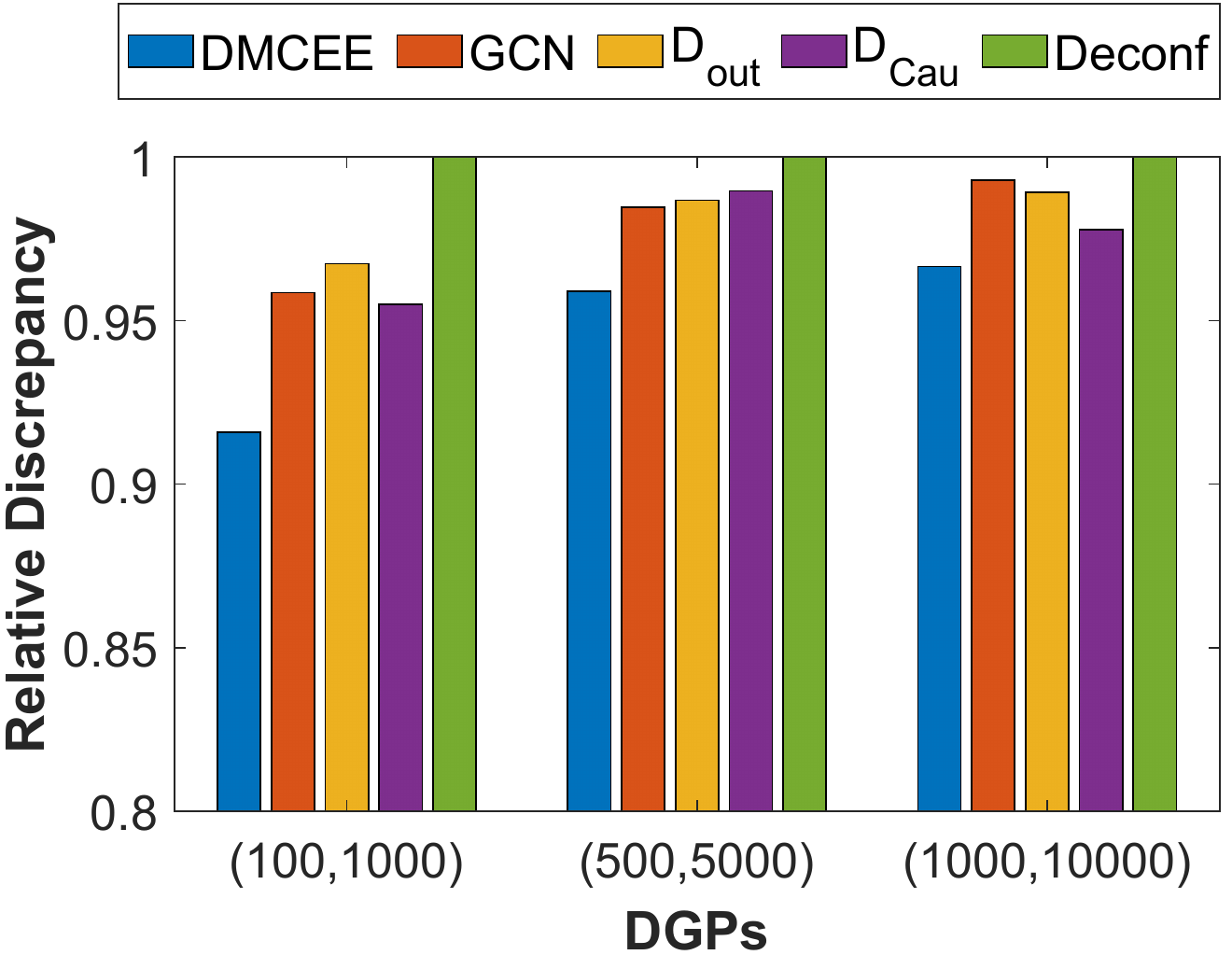}
  \caption{Comparisons of discrepancies between $\bm{\gamma}_R$ and $\mat{Z}$ among different models for all DGPs.}
 \label{confounder}
\end{figure}

We further examine the quality of the inferred HC. Specifically, we calculate the discrepancies between the estimated ($\bm{\gamma}_R$) and true HC ($\mat{Z}$), measured by Frobenius norm, i.e., $\|\bm{\gamma}_R-\mat{Z}\|_F$. We show the normalized results in Fig. \ref{confounder}. We observe that DMCEE accomplishes the most accurate estimations of HC for all DGPs and the improvement over the Deconfounder is significant. In fact, all models that approximate HC using the interaction graph $\mat{G}$ outperform Deconfounder w.r.t. the quality of inferred HC. This result suggests that network information is important for hidden confounding adjustment. The resulting high-quality surrogate confounder further justifies the advantages of learning representation of multi-modal proxies using the proxies encoding network and the causal adjustment network.
\subsection{Experiments on Yelp Online Reviews}
In this section, we examine the model utility on real-world datasets collected from Yelp and Google Map\footnote{https://www.google.com/maps}. As the ground-truth causal effects are unknown, we use a common evaluation method in causal inference with observational data \cite{liang2016causal,bonner2018causal}: \textit{OOD prediction} where the training and test data are from different distributions. 
\subsubsection{Data Descriptions}
Following \cite{luca2016reviews}, we curate novel datasets by merging two independent data sources for online reviews and restaurant popularity, respectively. The first data source\footnote{https://www.yelp.com/dataset/challenge} is the Yelp.com. It includes information of consumers and businesses in 10 metropolitan areas across 2 countries (USA and Canada). We filtered out non-restaurant businesses based on their category descriptions and selected two cities with the largest number of restaurants\footnote{We used a subset of original Yelp reviews due to the Google API limits and financial considerations for acquiring data from the second data source as detailed below.}: Las Vegas, U.S. and Toronto, Canada. 
\begin{table}
\small
\setlength\tabcolsep{2pt}
\begin{center}
\caption{Basic statistics of the \textit{LV} and \textit{Toronto} datasets. }
\begin{tabular}{ c|c|c|c|c } \hline
Dataset & \#Restaurants &\#Consumers &\# Ave. Reviews &Ave. Popularity \\ \hline
\textit{LV}&3,041 &304,102 & 255 (3--8,570)& 21.80 (0--100)\\\hline
\textit{Toronto}&3,828&68,026&67 (3--2,177)& 17.05 (0--100)\\ \hline
\end{tabular}
\label{review}
\end{center}
\end{table}
The second data source is Google Map that records hourly average consumer flow (00:00 AM--23:00 PM) of a restaurant from Monday to Sunday, i.e., \textit{popular times}. The value of popularity is on a scale of 0-100 with 1 and 100 denoting the least and most popular. 0 indicates the restaurant is closed during that hour. We select popularity of Saturday 07:00 PM - 08:00 PM as the outcome and remove restaurants closed during this period. We augmented the \textit{LV} and \textit{Toronto} datasets with the popularity dataset by matching restaurants' names and locations (a tuple of longitude and altitude). When this method failed or generated duplicate merges, we manually checked for the correct merge. This results in two complete datasets \textit{LV} and \textit{Toronto} that include both online reviews and restaurant popularity\footnote{Data can be downloaded at \url{https://github.com/GitHubLuCheng/Effects-of-Multi-Aspect-Online-Reviews-with-Unobserved-Confounders}}. Detailed analyses of these two datasets can be found in our previous work \cite{cheng2021effects}. Basic statistics of these two datasets can be seen in Table \ref{review}. We follow standard procedures in multi-aspect sentiment analysis to extract MAS from textual reviews in \textit{LV} and \textit{Toronto} datasets, as detailed in \cite{cheng2021effects}. The results include positive and negative sentiment scores w.r.t. five common aspects \cite{lu2011multi}: Food, Service, Price, Ambience and Anecdotal/Miscellaneous.
\subsubsection{OOD Popularity Prediction}
In OOD prediction, models that capture the causal relationships are expected to achieve better performance. To generate training and test data following different distributions, we group \textit{LV} and \textit{Toronto} datasets based on the observed covariates of restaurants (e.g., location, rating) using $k$-means algorithm, and compute the distance between the pair-wise cluster centroids $(\mat{c}_i,\mat{c}_j)$. We then take the cluster ($i$) that is furthest from the rest clusters as the test set and the rest as training set. Distance between the training and test data distributions are defined as:
\begin{equation}
    dist_{i}=\frac{\sum_{j\neq i}d(\mat{c}_i,\mat{c}_j)}{k-1} \quad \forall i\in\{1,...,k\},
\end{equation}
where $d(\cdot)$ denotes the Euclidean distance between two centriods. We iterate through all the clusters and the output $\mat{v}_{dist}\in \mathbb{R}^k$ is a vector with the $i$-th element equal to $dist_i$.
\begin{table}
\small
\setlength\tabcolsep{1pt}
\caption{OOD prediction using \textit{LV} dataset.}
\begin{subtable}[h]{\columnwidth}
\caption{MSE for predicting popularity using \textit{LV} dataset.}
\begin{center}
\begin{tabular}{c|c|c|c|c|c|c|c}\Xhline{1pt}
Hidden Covariate&$LR$&$LR_{con}$&$GCN$&$Deconf$&$D_{Out}$&$D_{Cau}$&$DMCEE$\\\Xhline{1pt}
$Location$&706.99 & \textbf{653.62} & 705.22 & 707.57 & 776.14 & 696.48& \underline{670.32}\\
$Category$&706.99 & \underline{661.77} &702.44 & 707.83&727.68& 690.90&  \textbf{656.28}\\
$Rating$&706.99&\underline{ 664.15} & 674.25 & 706.72 & 704.92 & 683.90 & \textbf{662.60} \\
\textit{Price Range}&706.99 & 710.25 & \underline{667.09}& 708.21 & 671.60 & 692.97 &  \textbf{664.39}\\\Xhline{1pt}
\end{tabular}
\end{center}
\end{subtable}
\begin{subtable}[h]{\columnwidth}
\caption{MAE for predicting popularity using \textit{LV} dataset.}
\begin{center}
\begin{tabular}{c|c|c|c|c|c|c|c}\Xhline{1pt}
Hidden Covariate&$LR$&$LR_{con}$&$GCN$&$Deconf$&$D_{Out}$&$D_{Cau}$&$DMCEE$\\\Xhline{1pt}
$Location$&22.53 & \textbf{21.47} &22.16&22.45 & 23.46& 22.32&\underline{21.58}\\
$Category$&22.53 & \underline{21.57} &22.39 & 22.59&22.66& 22.23&  \textbf{21.56}\\
$Rating$&22.53&\underline{21.72} & \underline{21.72} & 22.48&22.56 & 22.10 & \textbf{21.69} \\
\textit{Price Range}&22.53 &22.60 & \underline{21.62}& 22.42&22.05 & 22.16 &  \textbf{21.60}\\\Xhline{1pt}
\end{tabular}
\end{center}
\end{subtable}
\label{yelp1}
\end{table}
\begin{table}
\small
\setlength\tabcolsep{1pt}
\caption{OOD prediction using \textit{Toronto} dataset.}
\begin{subtable}[h]{\columnwidth}
\caption{MSE for predicting popularity using \textit{Toronto} dataset.}
\begin{center}
\begin{tabular}{c|c|c|c|c|c|c|c}\Xhline{1pt}
Hidden Covariate&$LR$&$LR_{con}$&$GCN$&$Deconf$&$D_{Out}$&$D_{Cau}$&$DMCEE$\\\Xhline{1pt}
$Location$&670.95 & \underline{630.04} & 661.51 & 673.72& 661.56& 681.14 & \textbf{612.13}\\
$Category$&670.95& \underline{635.45}&650.36& 671.26& 640.84& 685.06&  \textbf{623.73}\\
$Rating$&670.95 & \underline{626.52} &  655.00 &678.49& 635.15 & 677.56& \textbf{603.81}\\
\textit{Price Range}&670.95 & 676.53&641.51 & 676.17 & \underline{627.26} & 685.06&  \textbf{601.67}\\\Xhline{1pt}
\end{tabular}
\end{center}
\end{subtable}
\begin{subtable}[h]{\columnwidth}
\caption{MAE for predicting popularity using \textit{Toronto} dataset.}
\begin{center}
\begin{tabular}{c|c|c|c|c|c|c|c}\Xhline{1pt}
Hidden Covariate&$LR$&$LR_{con}$&$GCN$&$Deconf$&$D_{Out}$&$D_{Cau}$&$DMCEE$\\\Xhline{1pt}
$Location$&21.12 & \underline{20.34} & 21.16 & 21.19& 21.08& 21.29 & \textbf{20.24}\\
$Category$&21.12& \textbf{20.21}&20.81& 21.13& 20.70&21.34&  \underline{20.42}\\
$Rating$&21.12& \underline{20.20} & 21.12&21.19& 20.54& 21.26& \textbf{20.09}\\
\textit{Price Range}&21.12 & 21.31&20.74&21.11& \underline{20.59} & 21.34&  \textbf{20.06}\\\Xhline{1pt}
\end{tabular}
\end{center}
\end{subtable}
\label{yelp2}
\end{table}
We set $k=5$ for both datasets\footnote{Preliminary experiments suggested that $k$ almost has no influence on the results.}. The next step explicitly injects hidden confounding into the observational data: we hide one restaurant covariate (e.g., location) for both training and test data such that only the rest attributes can be ``observed''. Each restaurant has 4 covariates: \textit{Price Range, Location, Rating}, and \textit{Category}. This ends up with all in all 4 sets of experiments for each dataset. We report MAE and MSE on predicted outcomes in Table \ref{yelp1}-\ref{yelp2} with the best (boldface) and the second best results (underline) highlighted. We can observe that DMCEE and $LR_{con}$ mostly achieve the best and the second best performance w.r.t. both MAE and MSE. The improvement of DMCEE over Deconfounder w.r.t. MSE is significant, e.g., 7.3\% improvement using \textit{LV} with \textit{Category} being the hidden covariate. MSE is more sensitive to large errors compared to MAE, therefore, the performance differences are larger regarding MSE. Of particular notion is that $LR_{con}$ often outperforms \textit{GCN}. This finding suggests that directly using representation of multi-modal proxies without control of `bad' variables can induce undesired biases and the influence is large. DMCEE effectively excludes these biases through the inference models for causes and outcome in the causal adjustment network. The resulting representations are sufficient to control for confounding and are less biased to provide valid causal estimates for effects of MAS on popularity. 
\subsubsection{Case Studies with Dose-Response Curves}
To investigate the actionable insight of DMCEE on improving restaurant popularity, we further show how to use DMCEE to identify the aspects that restaurateurs should prioritize to improve popularity. In particular, we ask \textit{What will the restaurant popularity be if we improve the quality of an aspect to a certain level?} This experiment requires generating the \textit{dose-response curve} \cite{imbens2000role,schwab2019learning} with the $x$-axis being the dose, i.e., individual aspect sentiment score, and $y$-axis being the response, i.e., popularity. Intuitively, different types of restaurants are likely to be operated distinctively to improve different aspects. For example, improving the food quality of Fast Food typically has a larger influence on popularity compared to reducing the price. 

\begin{figure*}
\centering
\begin{subfigure}{0.25\textwidth}
\centering
  \includegraphics[width=.8\linewidth]{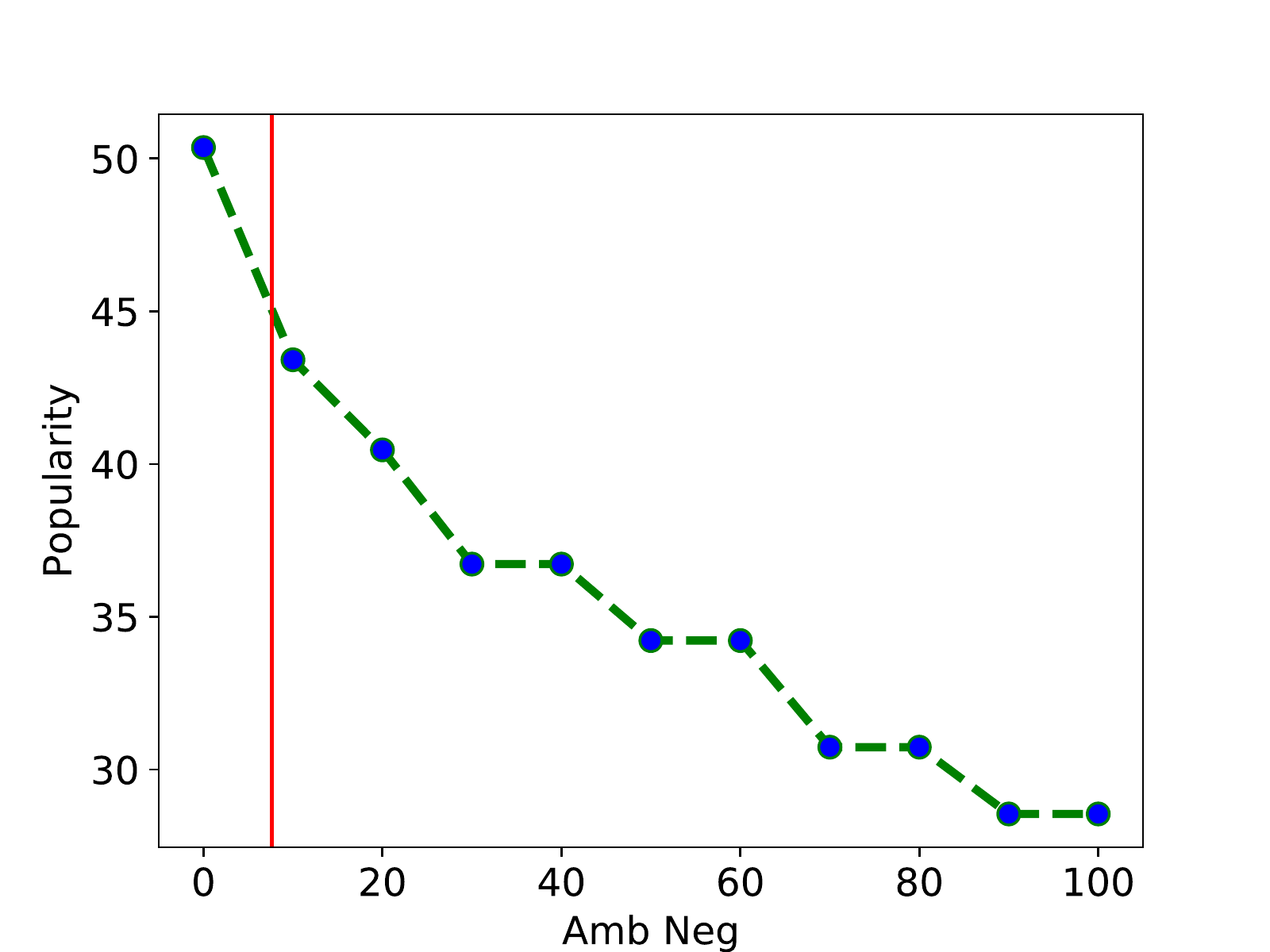}
  \caption{Bar: Ambience Neg.}
\end{subfigure}%
\begin{subfigure}{0.25\textwidth}
\centering
  \includegraphics[width=.8\linewidth]{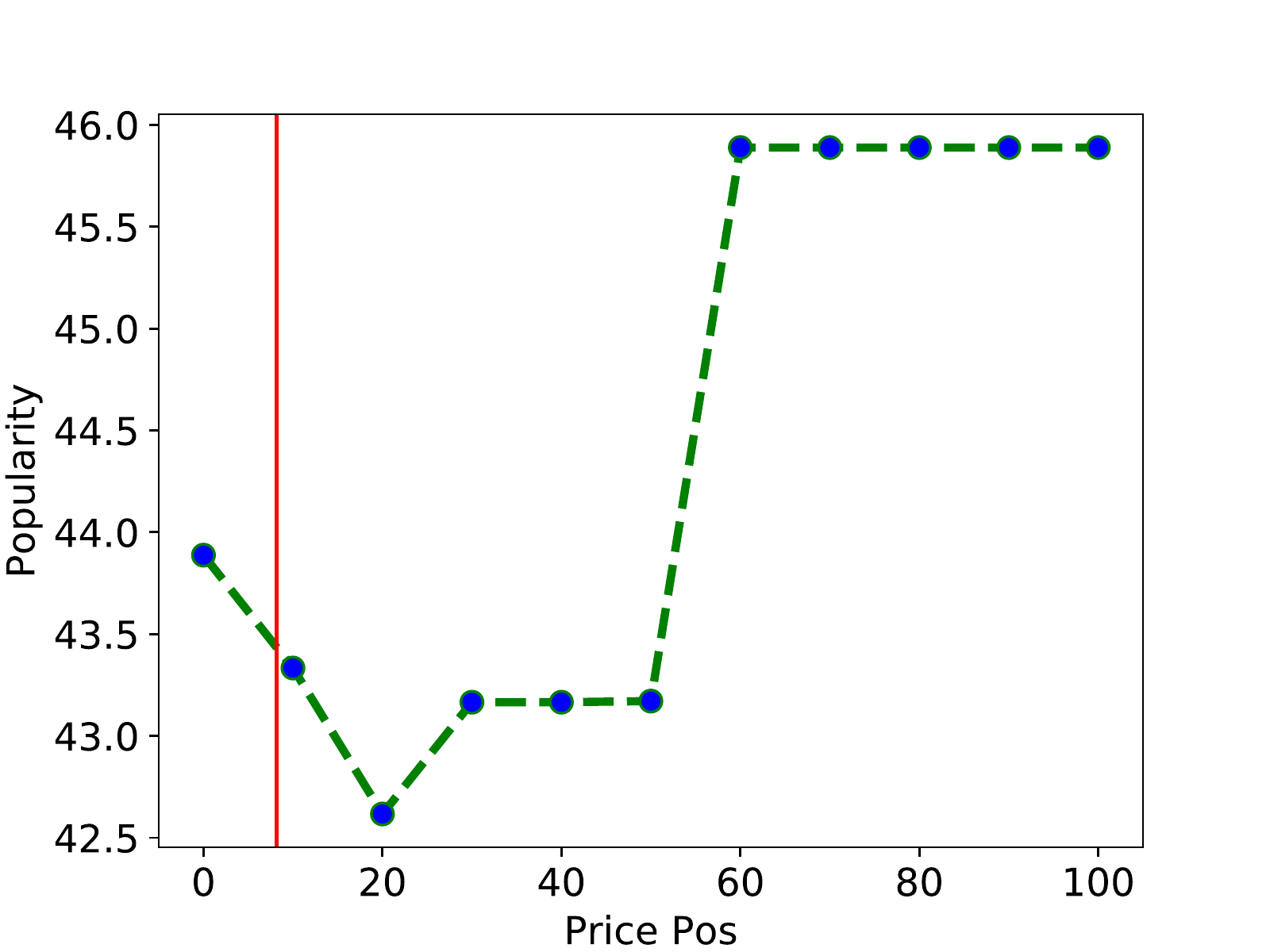}
  \caption{Bar: Price Pos.}
\end{subfigure}%
\begin{subfigure}{0.25\textwidth}
\centering
  \includegraphics[width=.8\linewidth]{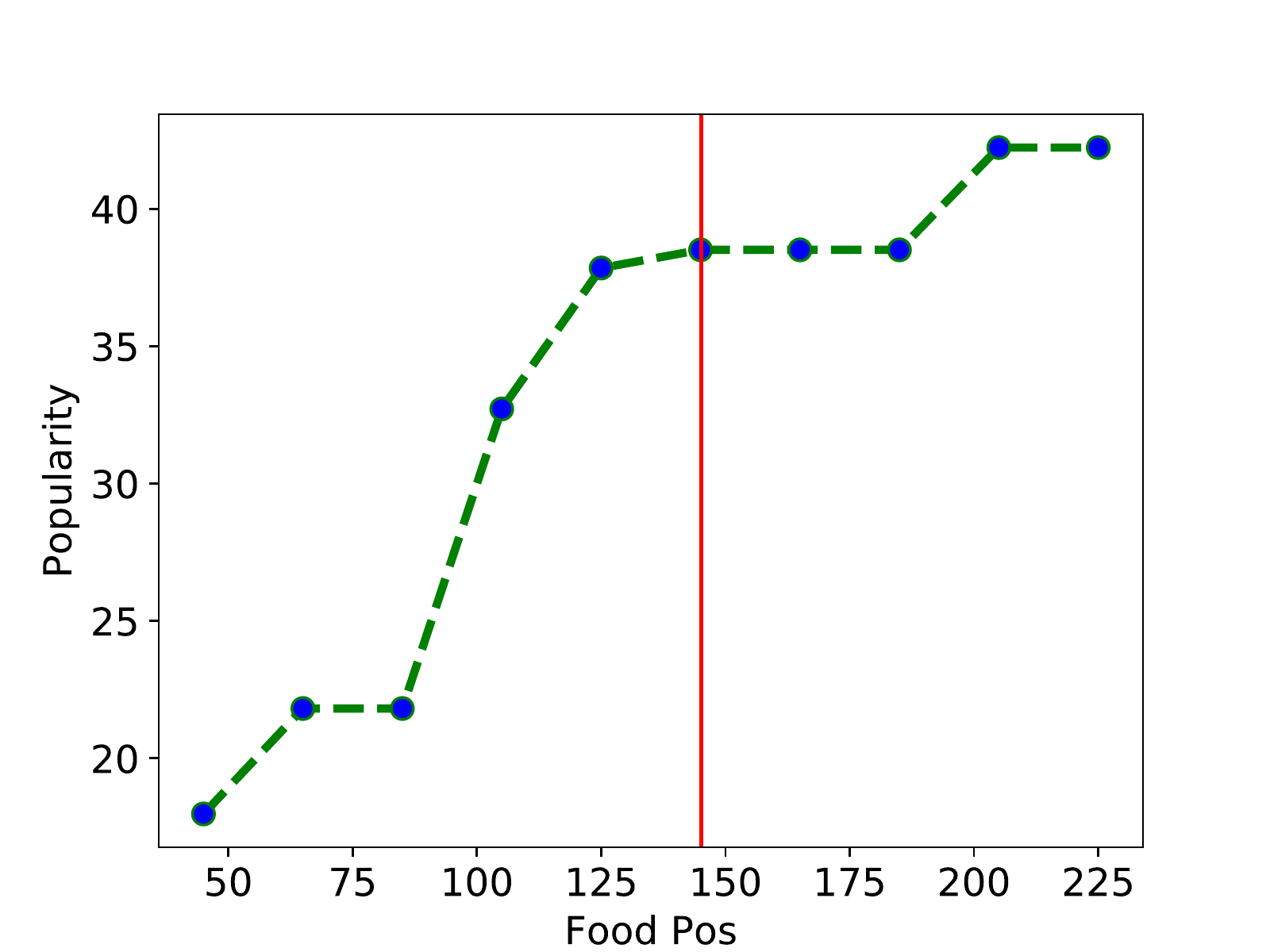}
  \caption{Bar: Food Pos.}
\end{subfigure}%
\begin{subfigure}{0.25\textwidth}
\centering
  \includegraphics[width=.8\linewidth]{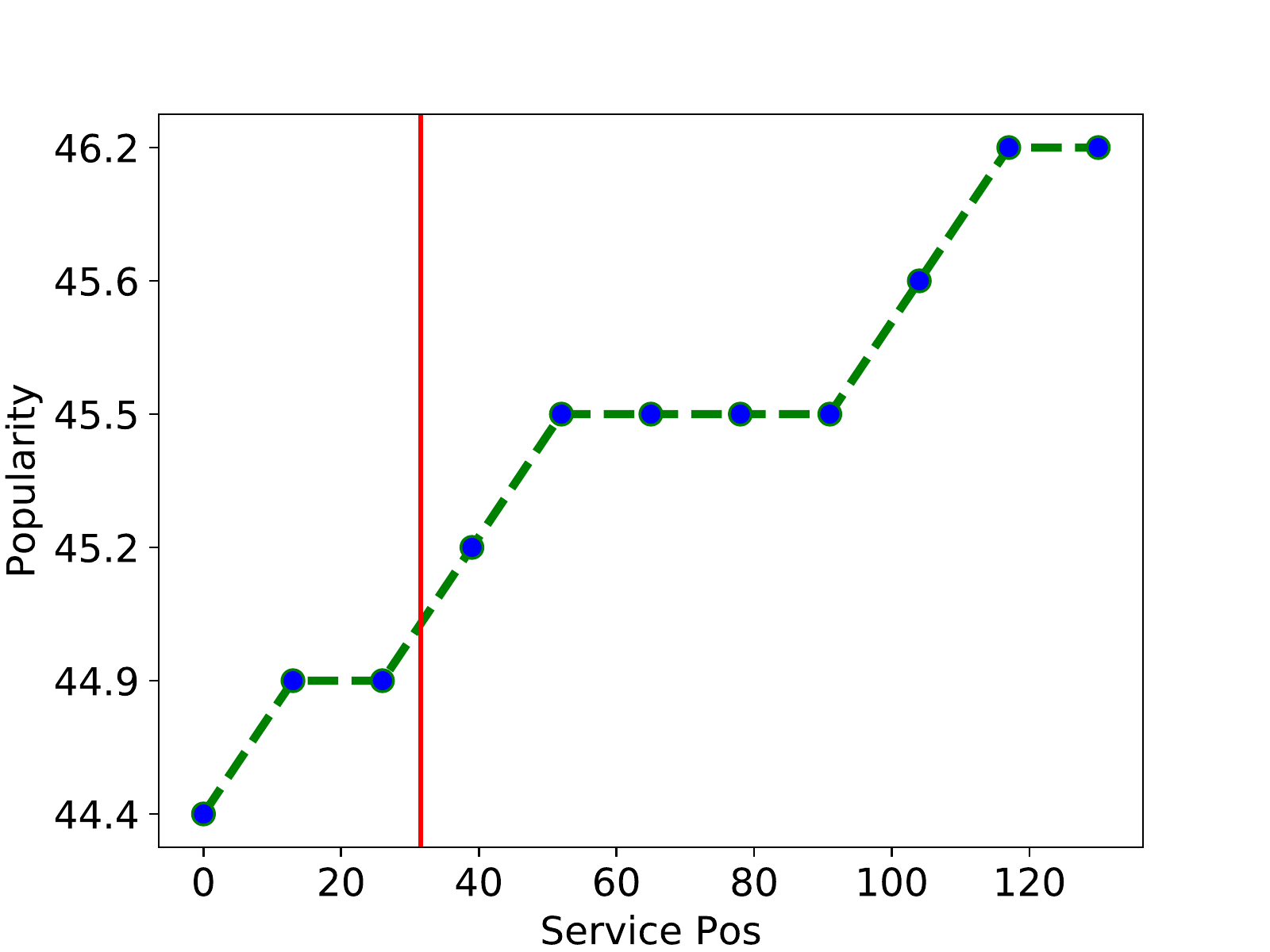}
  \caption{Bar: Service Pos.}
\end{subfigure}
\begin{subfigure}{0.25\textwidth}
\centering
  \includegraphics[width=.8\linewidth]{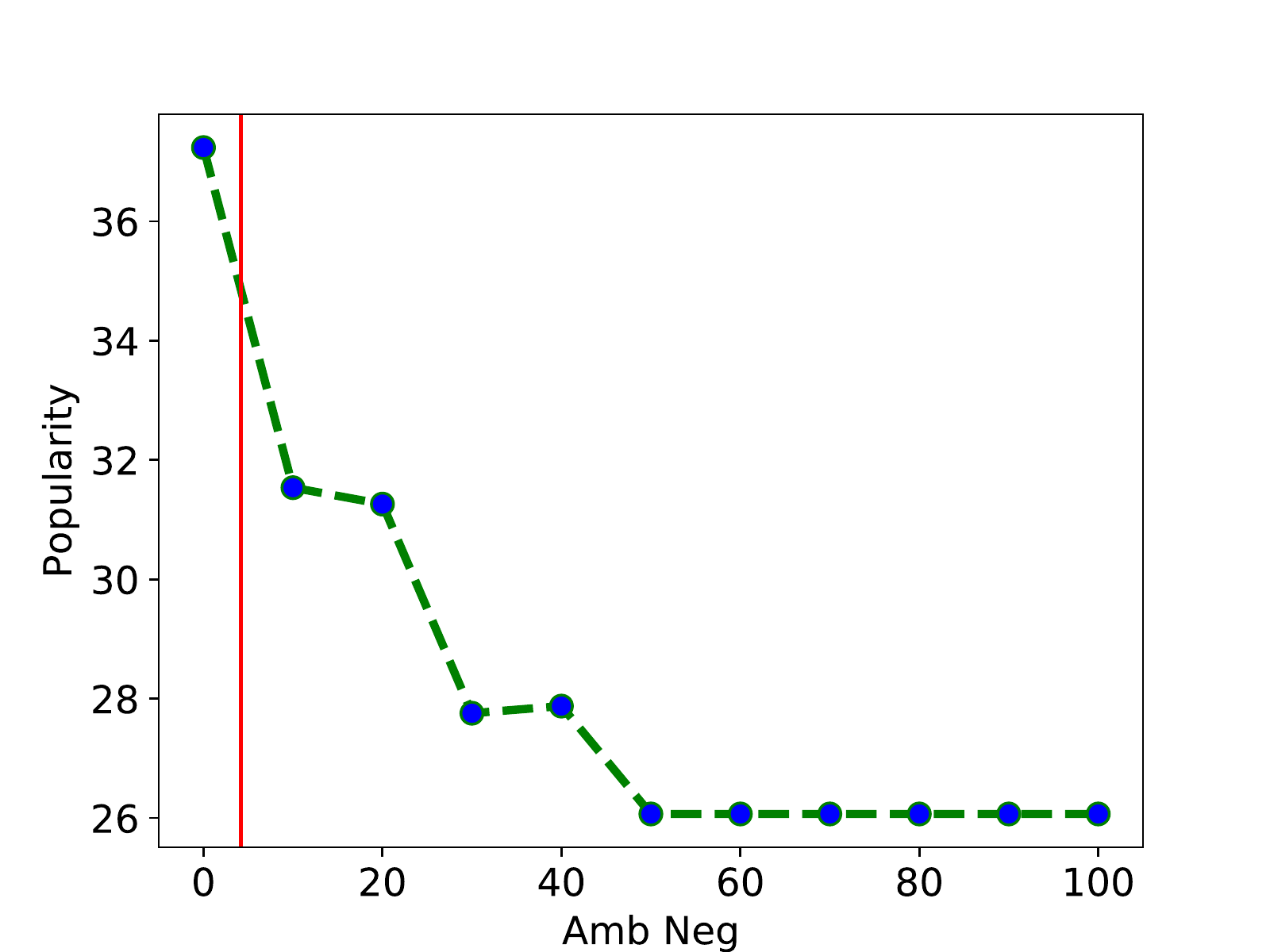}
  \caption{Fast: Ambience Neg.}
\end{subfigure}%
\begin{subfigure}{0.25\textwidth}
\centering
  \includegraphics[width=.8\linewidth]{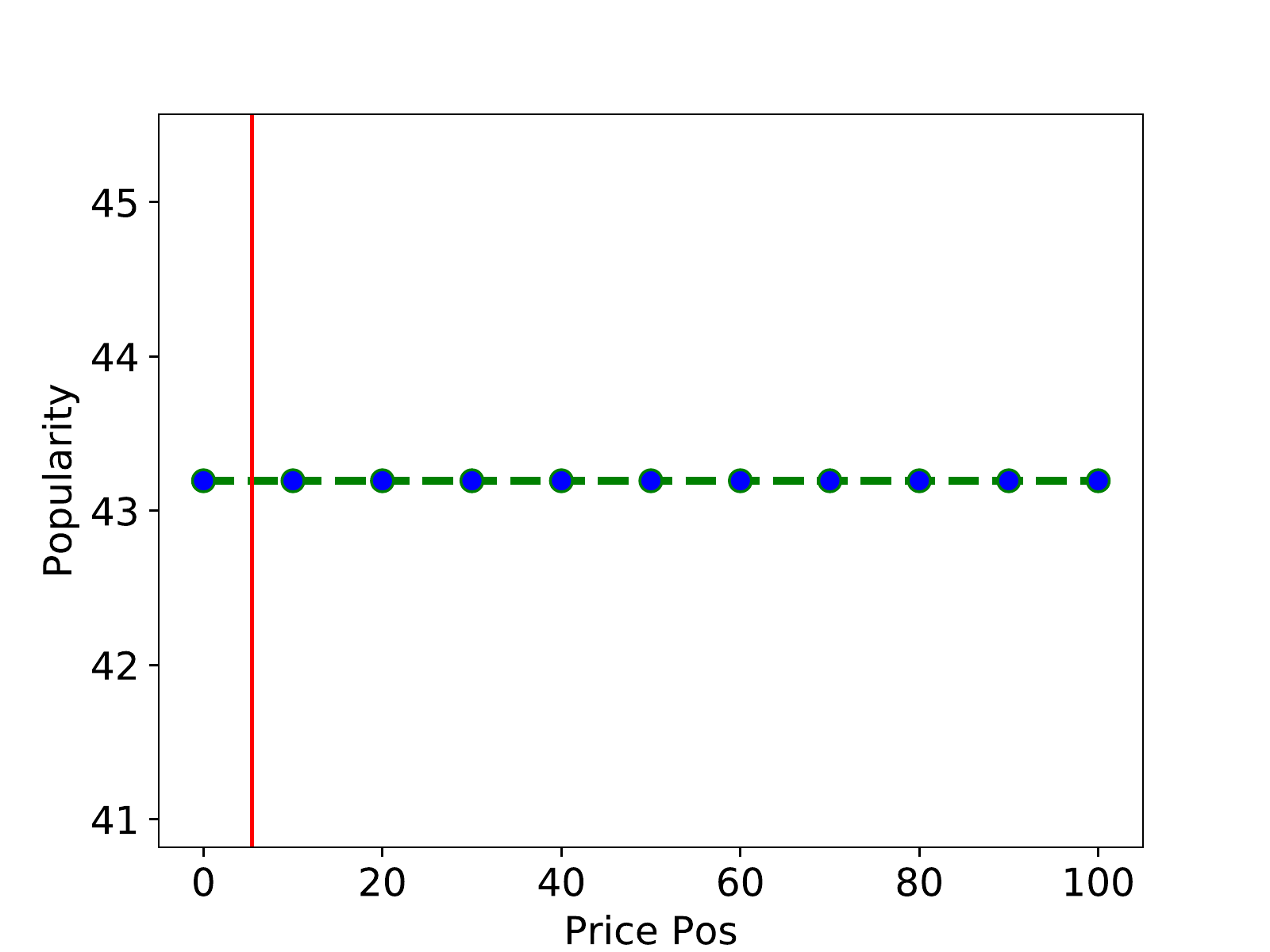}
  \caption{Fast: Price Pos.}
\end{subfigure}%
\begin{subfigure}{0.25\textwidth}
\centering
  \includegraphics[width=.8\linewidth]{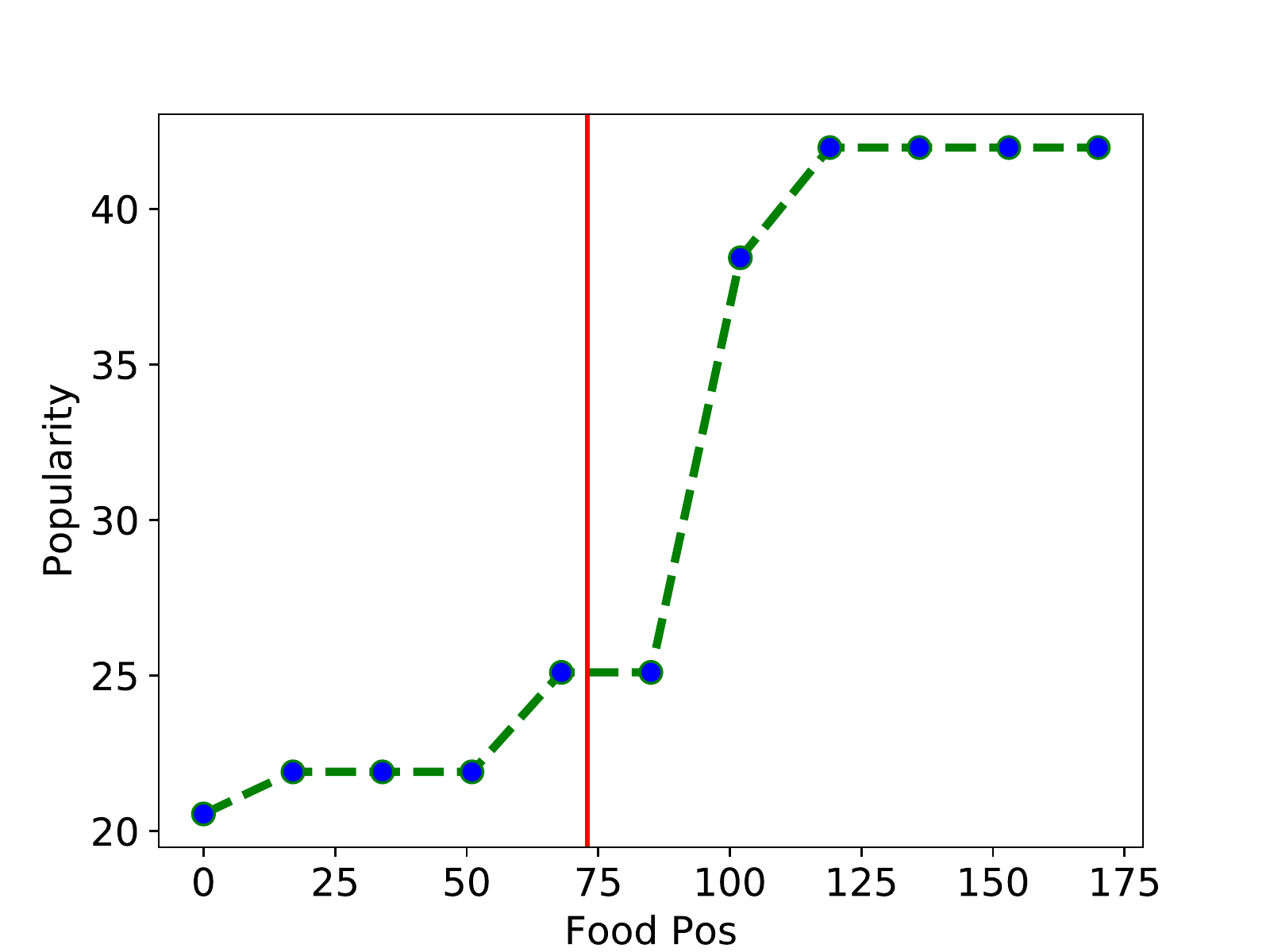}
  \caption{Fast: Food Pos.}
\end{subfigure}%
\begin{subfigure}{0.25\textwidth}
\centering
  \includegraphics[width=.8\linewidth]{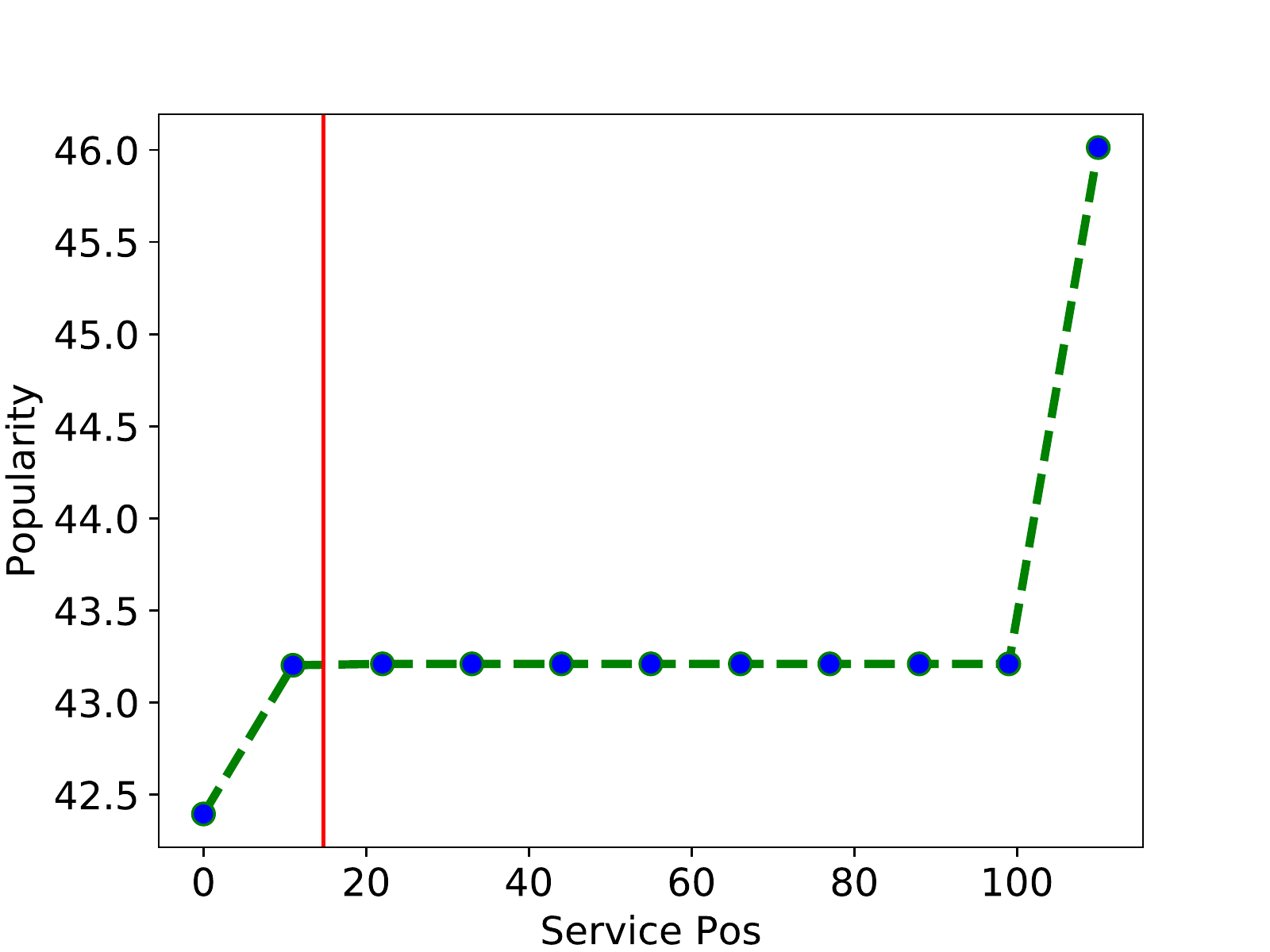}
  \caption{Fast: Service Pos.}
\end{subfigure}
\caption{Dose response curves for Bar and Fast Food (Fast) w.r.t. different sentiment aspects. The red line is the current aspect sentiment level.}
 \label{dose}
\end{figure*}
We first group restaurants into three categories based on their descriptions on Yelp: Bar, Fast Food, and Other. Given a group of restaurants with type $t\in\{\text{Bar, Fast Food, Other}\}$, we denote the estimated HC as $\mat{\hat{\gamma}}^t$ and popularity as $\mat{y}^t$. For sentiment aspect $j$ and restaurant type $t$, the dose-response curve is a function of sentiment level $l$ denoting the average of the estimated conditional expectation of the popularity \cite{imbens2000role}:
\begin{equation}
\small
    \overline{y}^t_{j}=\frac{1}{N_t}\sum_{i=1}^{N_t}f_j(\bm{a}_{i}^l,\mat{\hat{\gamma}}^t_{i}),
\end{equation}
where $N_t$ is the total number of restaurants with type $t$. $\bm{a}_{i}^l$ is the MAS of restaurant $i$ with sentiment level of aspect $j$ being $l$ while other aspect sentiment scores remain the same. $f_j(\cdot)$ denotes a non-linear model such as Random Forest. We then generate the response curve for sentiment aspect $j$ of restaurants with type $t$ by varying $l$. In the experiments, we define $f_j(\cdot)$ as the xgboost model \cite{chen2016xgboost} for its simplicity and superior performance. To generate more reliable results, we only plot dose-response curves for the aspects that have statistically significant (i.e., $p$-value $< 0.05$) causal effects on popularity. This experiment employs the \textit{Toronto} dataset and results for Bar and Other are similar, hence, we only present the response curves for Bar and Fast Food, as shown in Fig.  \ref{dose}.

\noindent\textbf{Implications.} \textit{What aspects should restaurateurs prioritize to effectively improve their restaurant popularity with limited budget?} Based on Fig. \ref{dose}(a), popularity of Bar can be effectively improved when the Ambience is better. An interesting observation from Fig.  \ref{dose}(b) is that a slight reduction of Price has limited influence on Bar popularity whereas a large reduction of Price will result in a substantial improvement of Bar popularity. From Fig. \ref{dose}(c), we find that improving the food quality might not be an effective strategy to increase Bar popularity. Rather, Bar owners can focus on Service as its influence on popularity is nearly linear as shown in Fig. \ref{dose}(d). For example, if increasing the current Bar service by 10\%, the popularity, on average, can approximately increase 0.1. Experiments on Fast Food present different results. For example, the influence of Ambience on popularity is significant when its quality is increased above a certain threshold as shown in Fig. \ref{dose}(e). Fig. \ref{dose}(f)-(g) delineate that although price is a major advantage of Fast Food, further reduction of price barely affects its popularity. However, the popularity can significantly grow via a slight improvement of food quality. Fig. \ref{dose}(h) reveals that improving Service of Fast Food is less effective.
\subsection{Parameter Analysis}
\begin{figure}
\centering
\begin{subfigure}{0.5\columnwidth}
\centering
  \includegraphics[width=\linewidth]{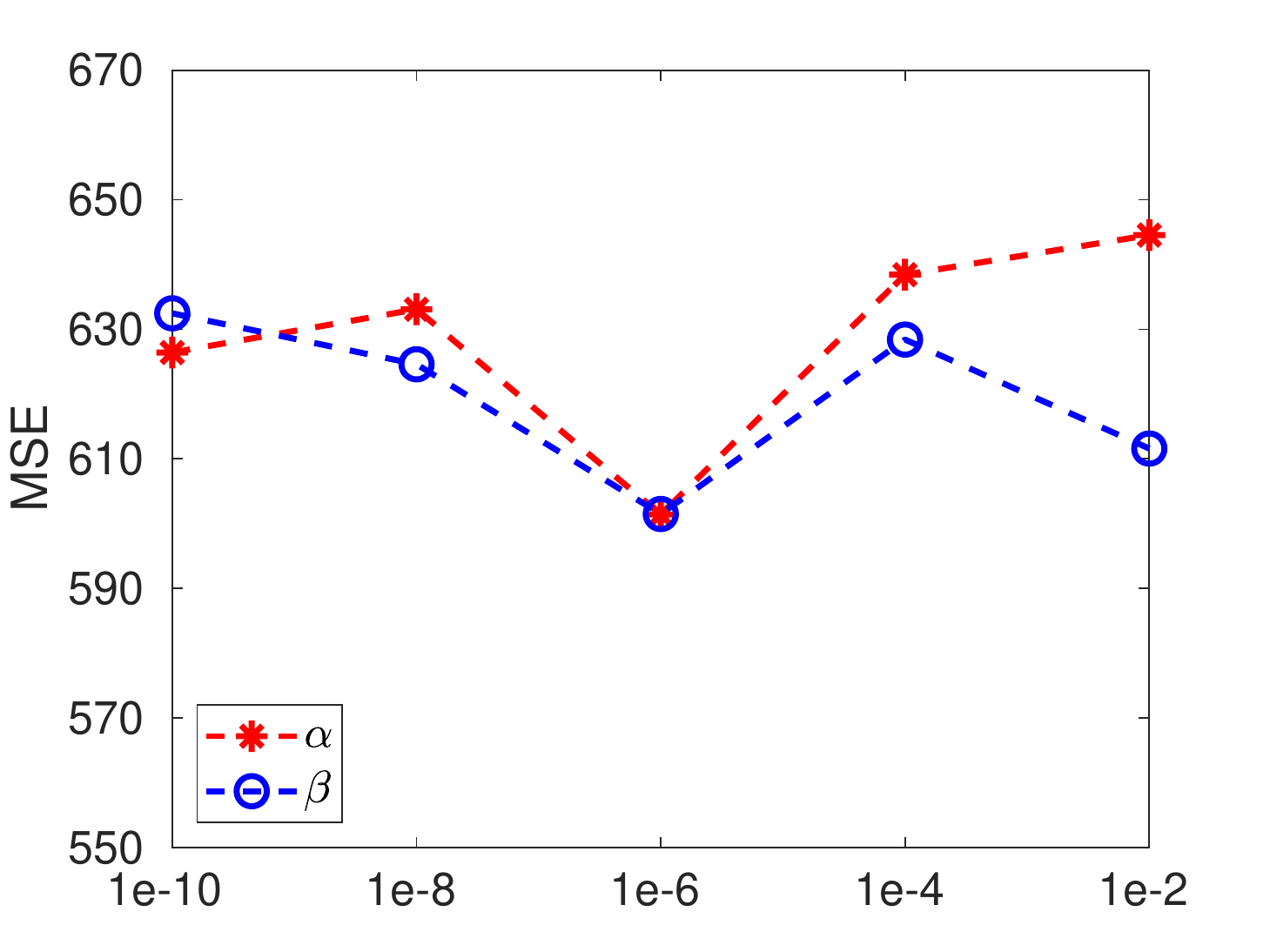}
  \caption{MSE.}
\end{subfigure}%
\begin{subfigure}{0.5\columnwidth}
\centering
  \includegraphics[width=\linewidth]{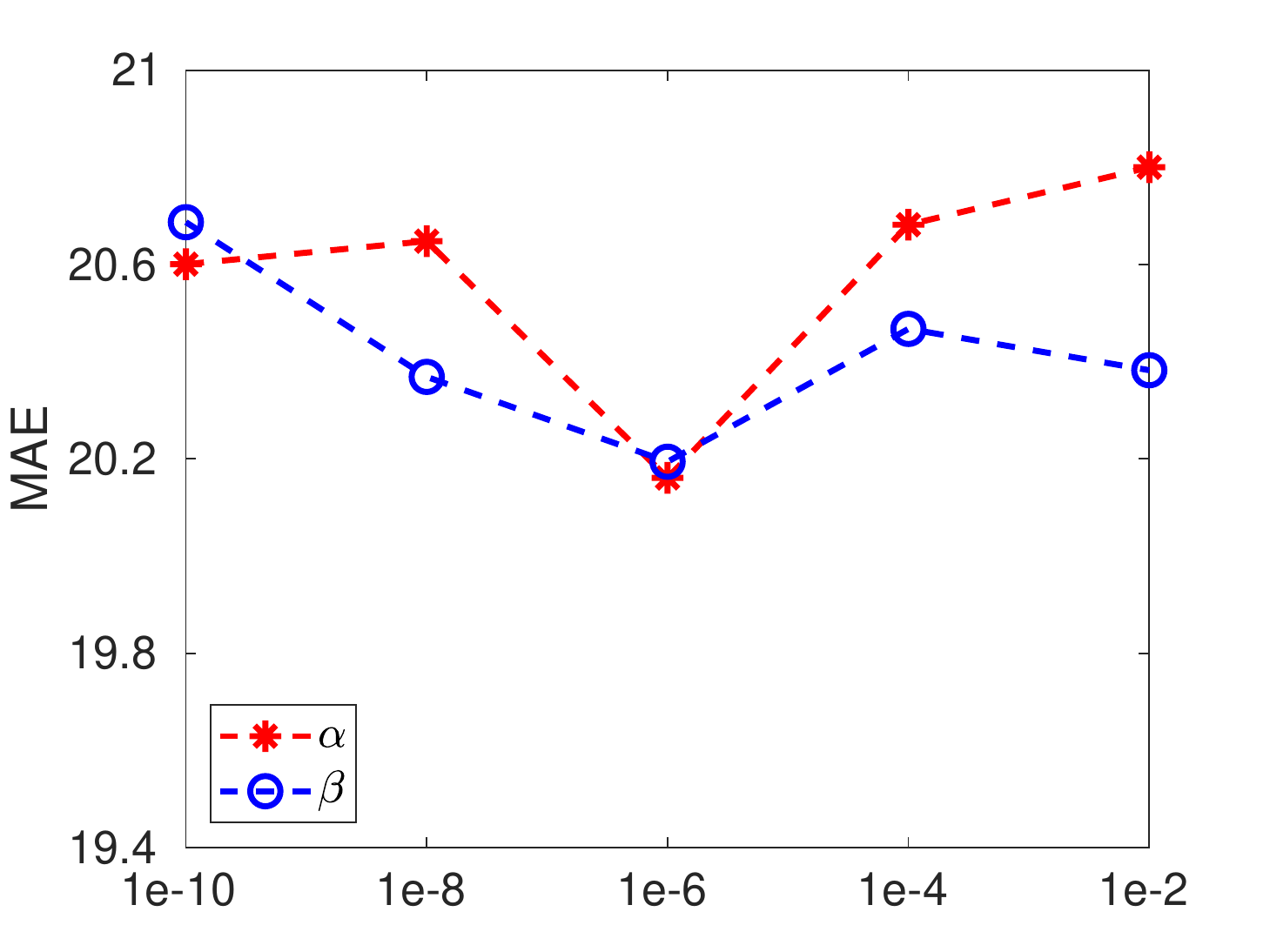}
  \caption{MAE.}
\end{subfigure}
\caption{Parameter analysis using \textit{Toronto} dataset.}
 \label{parameter}
\end{figure}

DMCEE has two main parameters: $\alpha$ and $\beta$ that together balance the contributions of the proxies encoding network and the causal adjustment network. This section investigates their effects on the model performance w.r.t. OOD prediction using \textit{Toronto} dataset. We vary the values of $\alpha$ and $\beta$ among $\{1\mathrm{e}{-}10,1\mathrm{e}{-}8,1\mathrm{e}{-}6,1\mathrm{e}{-}4,1\mathrm{e}{-}2\}$ and show the results of MSE and MAE in Fig.  \ref{parameter}. We can observe that DMCEE achieves the best MSE and MAE when both $\alpha$ and $\beta$ are around $1\mathrm{e}{-}6$. The performance of DMCEE tends to degrade as $\alpha$ increases or as $\beta$ decreases, that is, overemphasis of the proxies encoding network or de-emphasis of the causal adjustment network can exacerbate the model performance.
\section{Related Work}
\textbf{Causal Effect Estimation with Multiple Treatments.} 
Classical causal inference with multiple treatments typically uses generalized propensity scores (GPS) \cite{imbens2000role}. Methods that extend classical matching, subclassification, and weighting strategies to multiple treatments based on GPS can be seen in \cite{lechner2002program,zanutto2005using,rassen2013matching,yang2016propensity,lopez2017estimation,scotina2019matching}. Many established works rely on a strong assumption that observed covariates can fully account for confounding. The recently proposed \textit{Deconfounder} framework \cite{wang2019blessings} advocated to infer the substitute confounders by directly factorizing the multiple causes via latent variable models such as PCA \cite{ringner2008principal}. While Deconfounder opens up new research frontier in multiple causal inference in the presence of HC, its design suffers from inherent limitations \cite{d2019multi,ogburn2019comment,imai2019discussion}. Our work advances this line of research by advocating to use the representation of multi-modal proxies that is sufficient for confounding adjustment meanwhile excludes the potential biases induced by bad proxies. The benefits are mainly two folds: (1) multi-modal proxies provide multi-faceted views of HC, enabling more accurate and low-variance estimations of causal effects; (2) DMCEE considers the non-i.i.d. proxies, thereby advancing existing works in causal inference assuming i.i.d. proxy variables (e.g., \cite{miao2018identifying}).\\
\noindent\textbf{Causal Effect Estimation with Proxy Variables.} 
Proxy variable has been widely studied in causal inference with observational data \cite{nelson1991conditional,wooldridge2009estimating,louizos2017causal,miao2018identifying,pearl2012measurement}. Previous works have shown sufficient conditions for causal identification in the presence of HC using proxy variables, see, e.g., \cite{kuroki2014measurement,miao2018identifying,shi2018multiply}. In \cite{kuroki2014measurement,pearl2012measurement}, for example, the authors showed that the treatment effect can be identified with the method of effect restoration under several conditions. Weaker identification conditions proposed in \cite{miao2018identifying} require two independent proxy variables, one conditionally independent of causes given HC, and the other conditionally independent of outcome given HC. A recent work \cite{louizos2017causal} applied a variational autoencoder model to simultaneously approximate HC and estimate causal effects. However, most of these works assumed a single and binary cause. 
\\
\noindent\textbf{Causal Effect Estimation in Online Review Systems.} 
Prior empirical research on the effects of online reviews has established its importance in guiding consumer choices and business operations. For example, it has been shown that positive reviews, high ratings and popularity of reviews can largely increase book sales \cite{chevalier2006effect}, movie sales \cite{dellarocas2007exploring}, box office revenue \cite{reinstein2005influence}, sales of video games \cite{zhu2010impact}, and restaurant reservation availability \cite{anderson2012learning}. Based on user-generated online reviews, researchers also studied micro-level impact of documentary films on individual behavior \cite{rezapour2017classification}. Prior works, however, have been focused on a single and coarse-grained cause such as ratings without considering the granular information in textual reviews as well as the presence of HC. This research, however, investigates the effects of online reviews from multiple dimensions while controlling for hidden confounding.
\section{Conclusions}
This work studies \textit{causal effects} of multiple aspects in online reviews on business popularity. To account for HC in observational data, we propose to leverage the multi-modal proxies -- consumer and restaurant covariates and their interactions -- to learn a representation that is sufficient to account for hidden confounding meanwhile excludes potential biases induced by bad proxies. The resulting proxies encoding network and the causal adjustment network can jointly infer the surrogate confounder and estimate causal effects. Empirical evaluations on synthetic and real-world datasets corroborate the effectiveness of DMCEE. Our work can be extended to estimate individual treatment effect of each aspect to offer personalized operation strategies. We might also consider the popularity distribution as the outcome given that the effects of aspects on popularity during different time periods might be distinct. In the future, we can also consider the anchoring effect where previous reviews greatly affect future reviews due to data availability. 
\section*{Acknowledgements} This work is supported by ONR N00014-21-1-4002 and ARO  W911NF2110030. The views, opinions and/or findings expressed are the authors' and should not be interpreted as representing the official views or policies of the Army Research Office or the U.S. Government.
\bibliographystyle{ACM-Reference-Format}
\bibliography{sample-base}
\appendix
\section{Causal Identification of DMCEE}
With Pearl's $do$-calculus \cite{pearl2009causality}, causal identification ensures that an intervention distribution of the causes $P(Y|do(\mat{A}))$ is \textit{estimable} from the observed data. We describe in sketch that, under proper assumptions, effects of MAS on popularity are \textit{identifiable} following the causal mechanisms characterized in Fig. \ref{problem}, We use $\mat{X}$ to represent $\mat{X}_U$ and $\mat{X}_R$ to avoid repetition. Identification of $P(Y|do(\mat{A}))$ with proxies relies on two conditions \cite{miao2018identifying}: (1) at least two proxies of HC are observed and (2) one of the proxies is a null proxy that does not have impact on the outcome. Both conditions are satisfied under Fig. \ref{problem}. The following theorem is a direct extension from \cite{miao2018identifying}:
\begin{theorem}[Causal Identification of DMCEE] With the causal graph in Fig. \ref{problem} and the Asm. 2, $P(Y|do(\mat{A}))$ is identifiable:
\begin{equation}
    P(Y|do(\mat{A}))= \int P(Y|\mat{A},\mat{Z})P(\mat{Z})d\mat{Z}=\int h(Y,\mat{A},\mat{X})P(\mat{X})d\mat{X}
\end{equation}
for any solution $h$ to the integral equation
\begin{equation}
P(Y|\mat{A},\mat{G})=\int h(Y, \mat{A},\mat{X})P(\mat{X}|\mat{A},\mat{G})d\mat{X}.
\label{eqh}
\end{equation}
\end{theorem}
\noindent Thm. 2 enables DMCEE to formulate $ P(Y|do(\mat{A}))$ as a function of the observed data distribution $P(Y, \mat{A}, \mat{X}, \mat{G})$. We can therefore jointly identify the intervention distribution of individual sentiment aspect $P(Y|do(\mat{A}_{,j})),\ j\in\{1,2,...,2m\}$.
\section{Implementation}
Our proposed models were implemented in Python library Tensorflow \cite{abadi2016tensorflow}. Implementation code for baselines \textbf{GCN} and \textbf{Deconfounder} \cite{wang2019blessings} is adapted from \url{https://github.com/xiangwang1223/neural_graph_collaborative_filtering} and \url{https://colab.research.google.com/github/blei-lab/deconfounder_tutorial/blob/master/deconfounder_tutorial.ipynb}. We detail the parameter settings of the proposed models for both simulation data and the real-world datasets (i.e., \textit{LV} and \textit{Toronto}) in Table \ref{parameters}. The descriptions of the major parameters are introduced below:
\begin{itemize}
    \item Embed\_Size: the dimensions of consumer and restaurant embeddings. 
    \item Layer\_Size: the output size of every layer.
    \item $lr$: the learning rate.
    \item n\_layers: the number of hidden layers. 
    \item $\alpha$: the hyperparameter for the GCN module.
    \item $\beta$: the hyperparameter for the causes inference model.
    \item $\lambda$: the hyperparameter for $\ell_2$ regularization.
    \item Node\_Dropout: the keep probability w.r.t. node dropout for each deep layer.
    \item Mess\_Dropout: the keep probability w.r.t. message dropout for each deep layer.
\end{itemize}
\begin{table}
\caption{Details of the parameter settings in proposed models for both simulation data and real-world datasets.}
\begin{tabular}{|l|l|l|l|}
\hline
Parameter              & \textit{LV} & \textit{Toronto} & Simulation \\ \hline\hline
Epoch                  & 500          & 500 & 500        \\ \hline
Embed\_Size             & 32           & 32    & 15      \\ \hline
Layer\_Size            & 5           & 5     & 5    \\ \hline
Batch\_Size            & 1024         & 1024   & 32     \\ \hline
n\_layers              & 1            & 1   & 1       \\ \hline
$\alpha$ & 1e-6         & 1e-6 & 1e-6      \\ \hline
$\beta$ & 1e-6         & 1e-6  & 1e-6      \\ \hline
$\lambda$ & 1e-8         & 1e-8   & 1e-8      \\ \hline
$lr$       & 0.1         & 0.1   & 0.1       \\ \hline
Node\_Dropout          & 0.1          & 0.1  & 0.1         \\ \hline
Mess\_Dropout          & 0.1          & 0.1  & 0.1         \\ \hline
\end{tabular}
\label{parameters}
\end{table}
\end{document}